\pdfoutput=1
\documentclass[letterpaper]{article} 
\usepackage[draft]{aaai25}  
\usepackage{times}  
\usepackage{helvet}  
\usepackage{courier}  
\usepackage[hyphens]{url}  
\usepackage{graphicx} 
\usepackage{natbib}  
\usepackage{caption} 
\usepackage{algorithm}
\usepackage{algorithmic}

\usepackage{subcaption}

\usepackage{booktabs}
\usepackage{multirow}
\usepackage{amssymb}
\usepackage{amsmath} 

\usepackage{tcolorbox}
\tcbuselibrary{listings}
\tcbuselibrary{breakable}
\tcbuselibrary{skins}
\usepackage{multicol}
\usepackage{listings}
\usepackage{etoolbox}

\usepackage{fancyvrb}
\usepackage{framed}
\usepackage{verbatim}
\usepackage{fvextra}

\usepackage{hyperref}
\usepackage{CJKutf8}

\usepackage{pifont}
\usepackage{xcolor} 
\usepackage{xspace}
\usepackage{newfloat}
\usepackage{listings}

\newif\ifshowcomment
\showcommenttrue 

\newcommand{\vpara}[1]{\vspace{0.05in}\noindent \textbf{#1 }}

\newcommand{\model}{\texttt{BattleAgentBench}\xspace}

\newcommand\irule{\raisebox{-0.2em}{\includegraphics[width=1em]{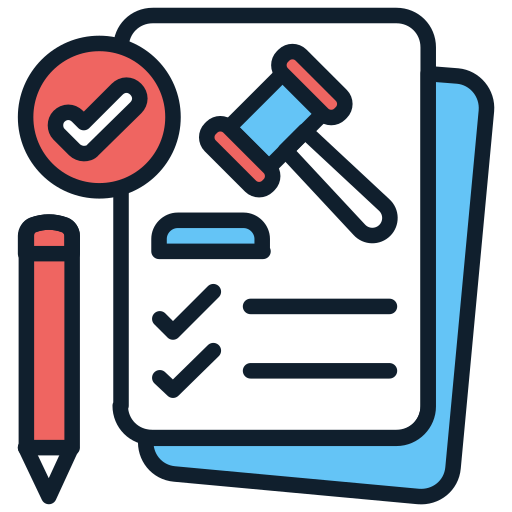}}}
\newcommand\ispace{\raisebox{-0.2em}{\includegraphics[width=1em]{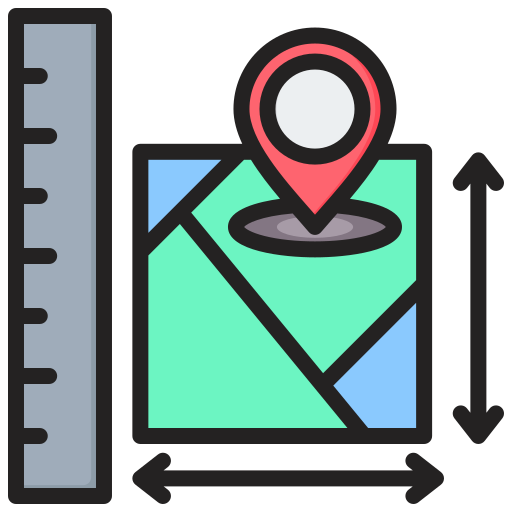}}}
\newcommand\igame{\raisebox{-0.2em}{\includegraphics[width=1em]{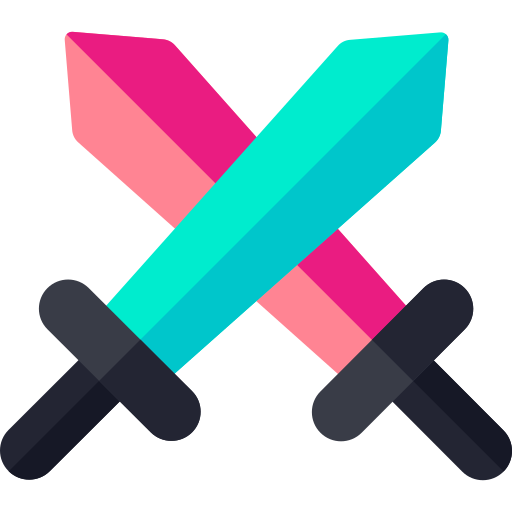}}}
\newcommand\isco{\raisebox{-0.2em}{\includegraphics[width=1em]{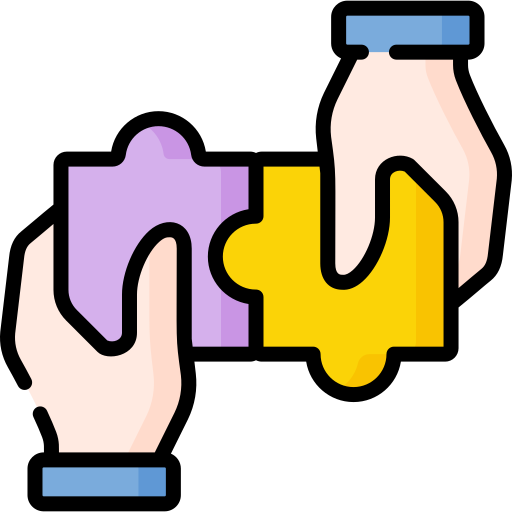}}}
\newcommand\idco{\raisebox{-0.2em}{\includegraphics[width=1em]{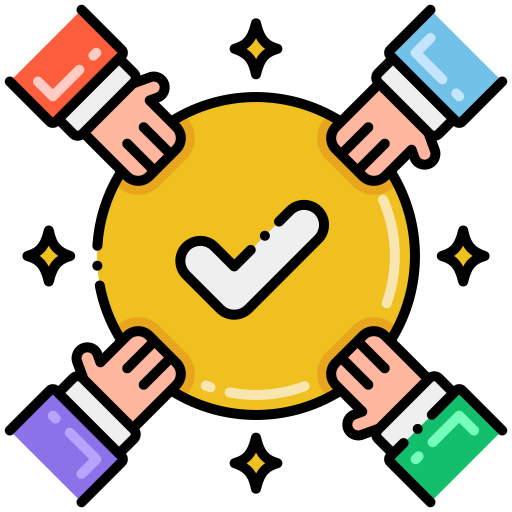}}}
\newcommand\iyes{\raisebox{-0.2em}{\includegraphics[width=1em]{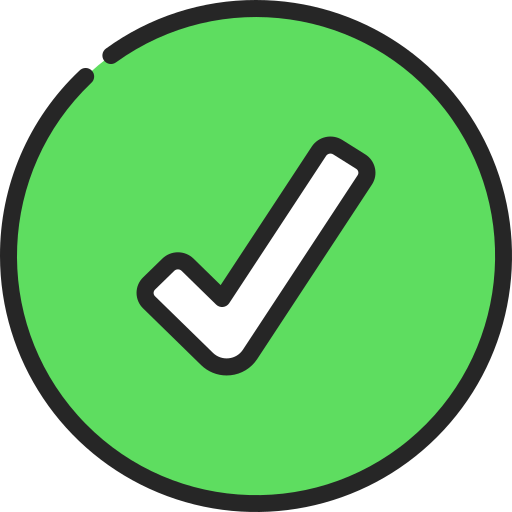}}}

\newcounter{myverbatimcounter}

\newtcblisting{myverbatim}[2][]{
  colback=gray!10,
  colframe=gray!50,
  boxrule=0.5pt,
  left=6pt,
  right=6pt,
  top=6pt,
  bottom=6pt,
  arc=0pt,
  boxsep=0pt,
  breakable,
  listing only,
  listing options={
    basicstyle=\ttfamily\small, 
    keepspaces=true,
    columns=flexible,
    breaklines=true,
    escapechar=|,
  },
  before upper*={\stepcounter{myverbatimcounter}%
    \par \vspace{-0.1pt}\noindent%
    \begin{minipage}{\linewidth}
      \centering
      \textbf{Figure \themyverbatimcounter: #2}
    \end{minipage}
  },
  #1
}

\newtcblisting{myverbatim3}[2][]{
  colback=gray!10,
  colframe=gray!50,
  boxrule=0.5pt,
  left=6pt,
  right=6pt,
  top=6pt,
  bottom=6pt,
  arc=0pt,
  boxsep=0pt,
  breakable,
  listing only,
  listing options={
    basicstyle=\ttfamily\small, 
    keepspaces=true,
    columns=flexible,
    breaklines=true,
    escapechar=|,
    numbers=none,
  },
  before upper*={\stepcounter{myverbatimcounter}%
    \par \vspace{-0.1pt}\noindent%
    \begin{minipage}{\linewidth}
      \centering
      \textbf{Figure \themyverbatimcounter: #2}
    \end{minipage}
  },
  #1
}

\DefineVerbatimEnvironment{myverbatim2}{Verbatim}{
  fontsize=\small,
  fontfamily=tt,
  commandchars=\\\{\},
  codes={\catcode`\$=3\catcode`\^=7},
  baselinestretch=1,
  xleftmargin=6pt,
  xrightmargin=6pt,
  frame=single,
  framesep=6pt,
  rulecolor=\color{gray!50},
  fillcolor=\color{gray!10},
}

\DeclareCaptionStyle{ruled}{labelfont=normalfont,labelsep=colon,strut=off} 
\floatstyle{ruled}
\newfloat{listing}{tb}{lst}{}
\floatname{listing}{Listing}
\title{\model: A Benchmark for Evaluating Cooperation and Competition Capabilities of Language Models in Multi-Agent Systems}
\author{
    Wei Wang, Dan Zhang, Tao Feng, Boyan Wang, Jie Tang
}
\affiliations{
    The Knowledge Engineering Group (KEG), Tsinghua University
}

\usepackage{bibentry}

\begin{document}

\maketitle

\begin{abstract}
Large Language Models (LLMs) are becoming increasingly powerful and capable of handling complex tasks, e.g., building single agents and multi-agent systems. 
Compared to single agents, multi-agent systems have higher requirements for the collaboration capabilities of language models.
Many benchmarks are proposed to evaluate their collaborative abilities.
However, these benchmarks lack fine-grained evaluations of LLM collaborative capabilities.
Additionally, multi-agent collaborative and competitive scenarios are ignored in existing works.
To address these two problems, we propose a benchmark, called \model, which defines seven sub-stages of three varying difficulty levels and conducts a fine-grained evaluation of language models in terms of single-agent scenario navigation capabilities, paired-agent task execution abilities, and multi-agent collaboration and competition capabilities.
We conducted extensive evaluations on leading four closed-source and seven open-source models. 
Experimental results indicate that API-based models perform excellently on simple tasks but open-source small models struggle with simple tasks. 
Regarding difficult tasks that require collaborative and competitive abilities, although API-based models have demonstrated some collaborative capabilities, there is still enormous room for improvement.
The code for \model is available at 
\url{https://github.com/THUDM/BattleAgentBench}.
\end{abstract}

\section{Introduction}

Large language models (LLMs) have showcased remarkable capabilities in handling intricate tasks~\cite{weather_forecasting, jumper2021highly, singhal2023large, zhang2024sciglm, zhang2024rest}, positioning them as pivotal agents in real-world scenarios such as gaming~\cite{R:10}, web automation~\cite{gur2023real}, and knowledge retrieval~\cite{wu2024avatar}. 
In particular, LLMs have been extensively employed to construct single-agent, double-agent, or multi-agent systems within gaming environments~\cite{R:8, R:9}. The solitary agent setups serve to assess the fundamental rule comprehension of LLMs, while multi-agent systems~\cite{R:4, R:5} demand heightened collaborative prowess. 
To gauge their collaborative aptitude effectively, various benchmarks~\cite{R:13, R:24, liu2023agentbench} have been introduced, albeit in a broad-stroke evaluation fashion. And one category of methods mainly focuses on collaborative scenarios~\cite{R:6, R:35}, while another category primarily focuses on competitive scenarios~\cite{R:8, R:9}.

However, appraising the collaborative proficiencies of multi-agent systems necessitates the definition of meticulous evaluation metrics and a profound understanding of their holistic skill sets. Moreover, multi-agent systems not only navigate collaborative landscapes but also navigate through complex competitive scenarios. For instance, consider the case of a badminton double challenge, wherein collaboration is imperative within an internal team while competitive dynamics unfold between two opposing teams. 
Nevertheless, these competitive scenarios often fall by the wayside in existing benchmarks due to the intricate evaluation tasks and metrics they entail.

\begin{table}[t!]
\resizebox{0.47\textwidth}{!}{%
\begin{tabular}{@{}c|cc|cc|ccc@{}}
\toprule
\multirow{1}{*}{Level} & \multicolumn{2}{c|}{Level 1} & \multicolumn{2}{c|}{Level 2} & \multicolumn{3}{c}{Level 3} \\ 
\cmidrule(l){1-8} 
\multirow{1}{*}{Ability/Stage}                         & 1               & 2              & 3         & 4         & 5        & 6                & 7               \\ \midrule
\irule \medspace Rule Understanding          &\iyes                 &\iyes                &\iyes           &\iyes           &\iyes          &\iyes                 &\iyes                \\
\ispace \medspace \medspace Spatial Perception       &\iyes                &\iyes               &\iyes           &\iyes           &\iyes          &\iyes                  &\iyes                 \\
\igame \medspace Competition             &                 &                &           &\iyes          &\iyes          &\iyes                  &\iyes                 \\
\isco \medspace Static Cooperation       &                 &                &\iyes           &           &\iyes          &                  &\iyes                  \\
\idco \medspace Dynamic Cooperation      &                 &                &           &           &         &\iyes                   &\iyes                  \\ \bottomrule
\end{tabular}
}
\caption{Evaluation abilities at each stage of each level of the \model.}
\label{tab:method_1}
\vspace{-5mm}
\end{table}

In light of these challenges, this paper introduces \model, a fine-grained benchmark to evaluate the collaborative and competitive capabilities of LLMs in multi-agent systems Specifically, as shown in Table~\ref{tab:method_1}, the \model framework encompasses three different difficult
y levels and seven stages.
At Level 1 (the single-agent level), we design two simple stages to assess a single agent's abilities of simple and complete game rule understanding. 
At Level 2 (the double-agent level), we use two stages with two agents to evaluate agents' cooperative and competitive task execution abilities. 
At Level 3 (the multi-agent level), we design three stages with different cooperative and competitive relationships between agents to assess agents' static cooperation competition, dynamic cooperation competition, and hybrid cooperation competition.

\begin{figure*}[th!]
    \centering
    \includegraphics[width=0.95\textwidth]{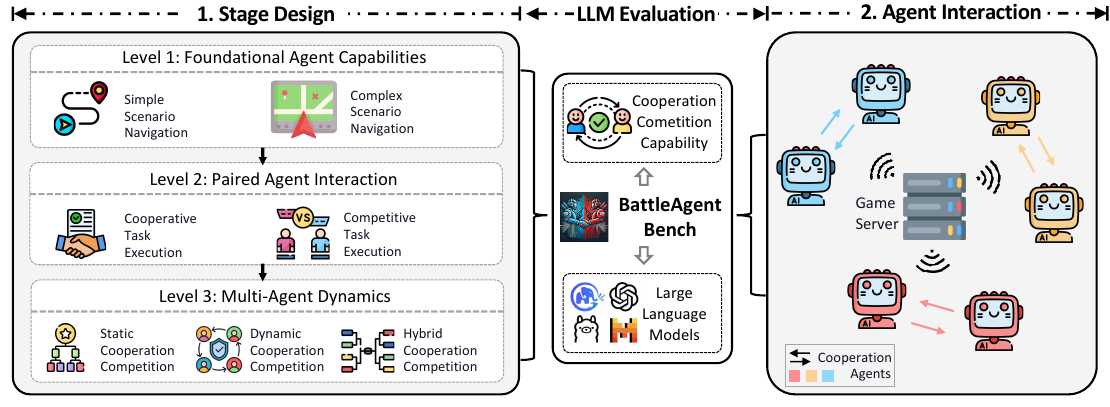}
    \caption{Overall evaluation framework of the \model.} 
    \label{fig:framework}
\end{figure*}

In addition, we have assessed 11 different LLMs using \model, including both four API-based and seven open-source models. We found that even in very simple stages, the performance of open-source small models was particularly poor, while API-based models performed well. However, in difficult tasks, although API-based models are still better than open-source small models, the gap between them has narrowed, and API-based models still have significant room for improvement.

In summary, the contributions of this work are:
\begin{itemize}
    \item We introduce the \model, a fine-grained multi-agent cooperation and competition evaluation benchmark, which includes three levels ranging from single agent to multiple agents.
    \item In the \model framework, we design seven stages from easy to difficult, capable of comprehensively assessing the basic abilities of single agents, cooperative abilities, and competitive abilities of multi-agents.
    \item We perform a thorough evaluation of 11 different LLMs using the \model, including the leading top four API-based LLMs and seven open-source models. Although API-based models perform well on simple stages, there is still significant room for improvement on complex stages.
\end{itemize}
\section{Preliminary}
We evaluate LLMs as agents in a turn-based interactive game, Battle City, where the agent takes corresponding actions based on the game state at each turn. Thus, the entire interaction process between the agent and the game can be viewed as a Markov Decision Process $(\mathcal{S}, \mathcal{A}, \mathcal{T}, \mathcal{R}, \mathcal{I}, \mathcal{O})$, which contains state space $\mathcal{S}$, action space $\mathcal{A}$, transition function $\mathcal{T}: \mathcal{S} \times \mathcal{A} \rightarrow \mathcal{S}$, reward function $\mathcal{R}$, game instruction $\mathcal{I}$, and observation space $\mathcal{O})$. The game server is responsible for state transitions and providing rewards. The agent needs to make an action $a$ based on instruction $i$, observation $o$, and state $s$. We apply LLMs to implement the agent's decision-making function:
\begin{equation}
    P(a_t|i,o_t,s_t) = \text{LLM}(i,o_t,s_t)
\end{equation}

\section{The \model}
To evaluate the cooperation and competition capabilities of large language models in a fine-grained manner, we design a new benchmark: \model. As shown in Figure \ref{fig:framework}, the \model consists of two main parts: stage design and agent interaction. In stage design, we design \textbf{seven stages of three different difficulty levels}. In agent interaction, we implement interactions between agents and servers to support evaluation in the above stages. Next, we introduce the overall evaluation framework and describe each stage in detail.

\subsection{Composition of \model}

\vpara{1. Stage Design.}
We design three different difficulty levels: Foundational Agent Capabilities (Level 1), Paired Agent Interaction (Level 2), and Multi-Agent Dynamics (Level 3), for a fine-grained evaluation of the agents' collaborative and competitive abilities.
At Level 1 (the single-agent level), we mainly evaluate a single agent's basic abilities, for example, whether can understand the game's basic rules. 
Regarding Level 2 (the double-agent level), we assess two agents' collaboration capabilities and competitive abilities in two stages. 
At Level 3 (the multi-agent level), we evaluate both the collaboration and competitive abilities of multiple agents in more complex scenarios. Each stage corresponds to a specific stage setting, defining the unique game environment for that stage, such as the number of players, the number of bases, the win/lose logic of the game, etc. 


\vpara{2. Agent Interaction.} 

\vpara{Game Server.} At the start of the game, the game server loads the specific environment according to the stage setting to be evaluated. During the game, the game server is responsible for sending observation information to agents, receiving actions 
from agents, updating the status of agents and environment, calculating agents' rewards, and determining whether the game has ended.

\vpara{LLMs as Game Agents.} Each agent represents a player, and its core function is the decision-making function, which determines the next action based on the received observation data. We use LLMs to implement each action function. The game observation data is converted into text format based on predefined templates.
Then, we prompt observation text to LLMs to obtain the output action.

\vpara{Cooperation between LLM Agents.} In addition to the interaction between agents and the server, we have also implemented a message communication interface between agents to support the evaluation of agents' static cooperation ability and dynamic cooperation ability.

\begin{figure}[t!]
    \centering
    \begin{subfigure}{0.23\textwidth}
        \centering
        \includegraphics[width=\textwidth]{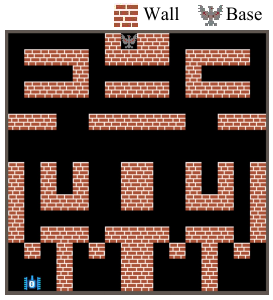}
        \caption{Level 1 - Stage 1}
        \label{fig:s1}
    \end{subfigure}
    \begin{subfigure}{0.23\textwidth}
        \centering
        \includegraphics[width=\textwidth]{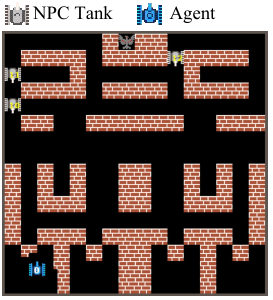}
        \caption{Level 1 - Stage 2}
        \label{fig:s3}
    \end{subfigure}
    \caption{Level 1: Stage 1 and Stage 2. The agent's goal in both stages is to reach the base location.}
    \label{fig:m_single}
\end{figure}

\subsection{Basic Rules of BattleAgent}
Specifically, the game contains four elements: the map, walls, bases, and tank as shown in Figure~\ref{fig:m_single}. Details and rules of this game are as follows:
\begin{itemize}
    \item The game map size is 512x512, with (0, 0) representing the top-left corner and (512, 512) representing the bottom-right corner.
    \item The map contains tanks, bases, and walls; the tank size and the base size are 32x32, and the wall size is 8x8.
    \item Tanks can move in four directions: up, down, left, and right. Walls and other tanks can obstruct the movement of the tank.
    \item The vertical movement range of the tank is from 0 to 512, and the horizontal movement range is 0$\sim$512.
    \item Tanks have four orientations: up, down, left, and right, and shooting can destroy the tank or wall in front of them in the current direction.
    \item When the tank's front faces the map boundary, it cannot continue to move forward.
    \item If there is a wall in front of the tank, it needs to shoot to eliminate the wall before it can continue to move forward.
\end{itemize}

\subsection{Level 1: Foundational Agent Capabilities}
Compared to directly evaluating the collaboration abilities of multiple agents in complex collaborative and competitive scenarios, we believe that it is necessary to first assess the task completion capabilities of individual agents within the environment. 
This approach allows us to decouple the assessment of the agents' basic capabilities from their collaborative abilities to some extent.
In complex environments, if an agent performs poorly, it becomes difficult to differentiate whether the subpar performance is due to the agent's basic capabilities or its collaborative abilities. Thus, we designed Level 1 (the single-agent level) to test basic capabilities in Figure~\ref{fig:m_single}.

\vpara{Stage 1: Simple Scenario Navigation.} In this stage, we test the agent’s understanding of the game rules and the spatial perception ability regarding simple scenario navigation. \textbf{a) Agent Setting:} There is only one agent and one base. \textbf{b) Agent Goal:} We start with a simple goal, which is to reach the base location as quickly as possible.
To achieve the goal, the agent first needs to understand the rules of the game with regard to these four elements.
In addition to understanding these rules, the agent must also have good two-dimensional spatial perception skills to move correctly toward the target.

\vpara{Stage 2: Complex Scenario Navigation.} We assess the ability to handle dynamic obstacles. \textbf{a) Agent Setting} and \textbf{b) Agent Goal} are consistent with Stage 1. The agent needs to eliminate the tanks that pose a threat to itself while approaching the base. \textbf{c) Change:} Compared to Stage 1, this stage introduces interfering Non-Player Character (NPC) tanks. These NPC tanks will move and shoot randomly. This requires the agent to have the ability to handle dynamic obstacles. 

\noindent In the next five stages, we also keep the basic game environment unchanged, primarily increasing the number of agents from different factions and setting more complex objectives.

\begin{figure}[t!]
    \centering
    \begin{subfigure}{0.23\textwidth}
        \centering
        \includegraphics[width=\textwidth]{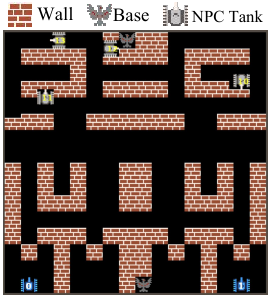}
        \caption{\small Level 2 - Stage 3}
        \label{fig:d1}
    \end{subfigure}
    \begin{subfigure}{0.23\textwidth}
        \centering
        \includegraphics[width=\textwidth]{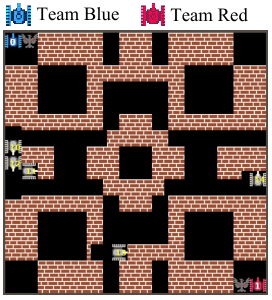}
        \caption{\small Level 2 - Stage 4}
        \label{fig:d2}
    \end{subfigure}
    \caption{Stages of Level 2 (double-agent level). In Stage 3 and Stage 4, the two agents have a cooperative relationship and a competitive relationship respectively.}
    \label{fig:m_double}
\end{figure}

\begin{figure*}[t!] 
\centering 


\begin{subfigure}{0.26\textwidth} \centering \includegraphics[width=\textwidth]{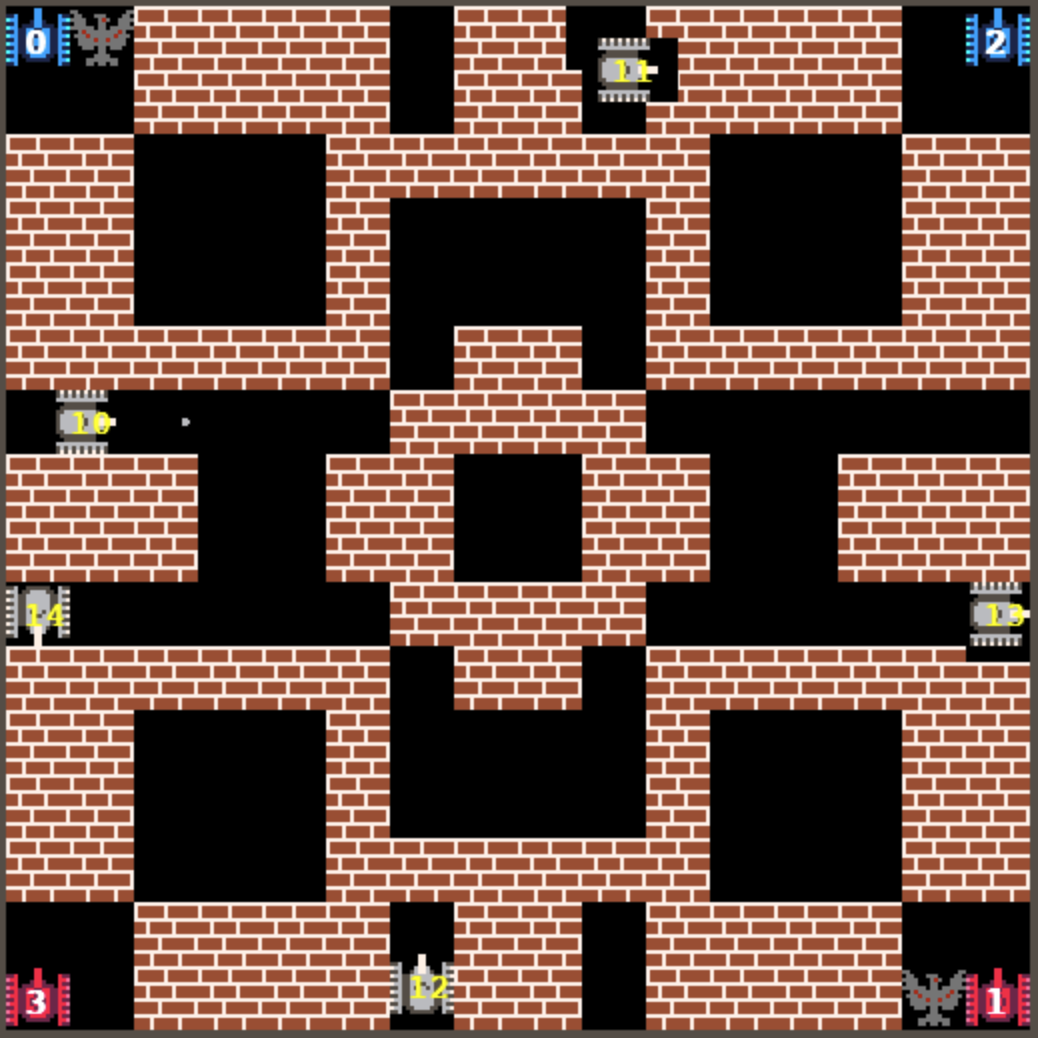} \caption{\small Level 3 - Stage 5} \label{fig:c3_5} \end{subfigure}
\hfill \begin{subfigure}{0.26\textwidth} \centering \includegraphics[width=\textwidth]{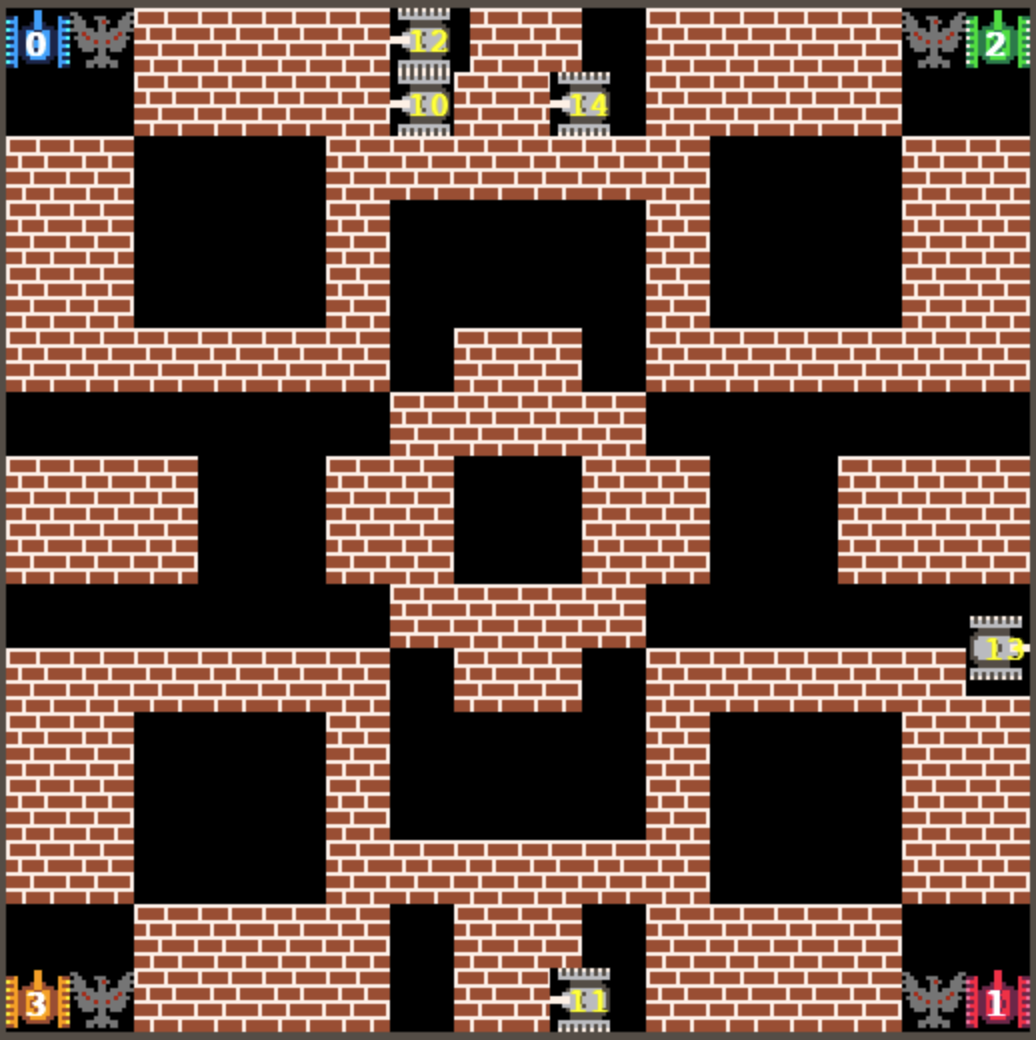} \caption{\small Level 3 - Stage 6} \label{fig:c3_6} \end{subfigure}%
\hfill \begin{subfigure}{0.26\textwidth} \centering \includegraphics[width=\textwidth]{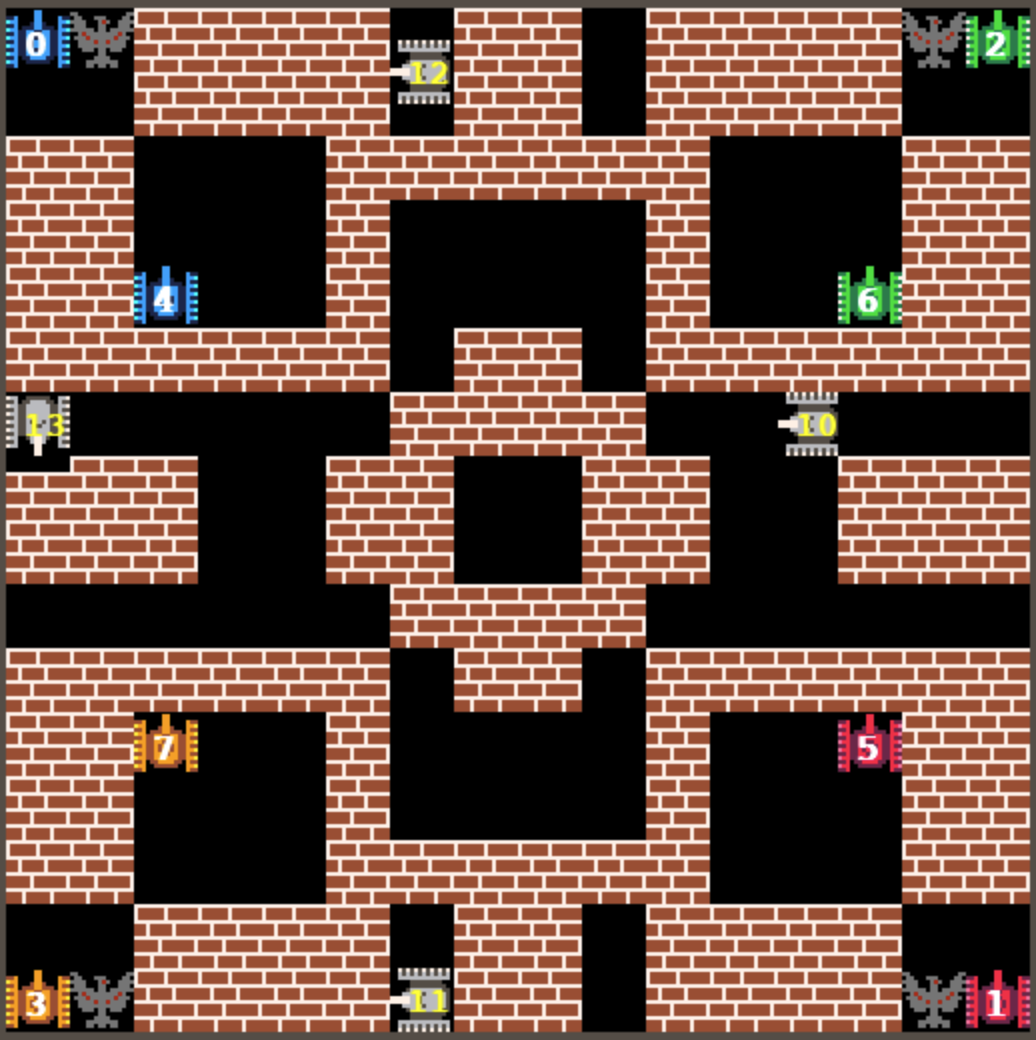} \caption{\small Level 3 - Stage 7} \label{fig:c3_7} \end{subfigure}
\hfill \begin{subfigure}{0.1\textwidth} \centering \vspace{-1.5em}
\includegraphics[width=\textwidth]{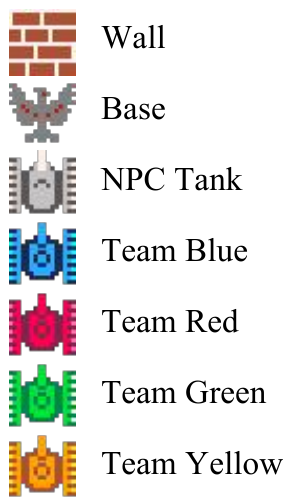} 
\label{fig:c3} \end{subfigure}
\caption{Stages of Level 3 (multi-agent level). In Stage 5, the agents within the team have a cooperative relationship, while the agents between teams have a competitive relationship. In Stage 6, the agents between teams have a competitive relationship, while allowing for cooperative relationships between teams. In Stage 7, the relationship within the team is cooperative, the relationship between teams is competitive, and cooperation between teams is also allowed.} 
\label{fig:combined2} 
\end{figure*}

\subsection{Level 2: Paired Agent Interaction}
Compared to Level 1 (single-agent level), Level 2 (double-agent level) introduces another agent, allowing for further evaluation of the agent's cooperation and competition abilities. Moreover, compared to a multi-agent level, the two-agent environment can assess the agent's cooperation and competition abilities with the minimum number of agents, minimizing the influence of the number of agents on the results. The collaborative relationships among multiple agents are more complex, so we designed this level to gradually increase the difficulty of assessing collaborative and competitive capabilities, as shown in Figure~\ref{fig:m_double}.

\vpara{Stage 3: Cooperative Task Execution.} In this stage, we evaluate the cooperative task execution ability of agents. \textbf{a) Agent Setting:} two agents belong to the same faction. There are two bases: one is the own base, and the other is the enemy base. \textbf{b) Agent Goal:} The goal of the two agents is to eliminate the enemy base while protecting their own base from being attacked. \textbf{c) Change:} The two agents have a cooperative relationship and need to work together to accomplish this objective. 

To facilitate cooperation between the agents, we have introduced a communication interface in the game that allows the two agents to send collaboration messages to each other in natural language. In each action step, an agent can autonomously decide whether to send a collaboration message and propose a specific cooperation plan. The other agent can then reply to the collaboration message, indicating whether it accepts or rejects the proposed cooperation plan.

\vpara{Stage 4: Competitive Task Execution.} In this stage, we assess the competitive task execution abilities of agents. \textbf{a) Agent Setting:} There are two agents and two bases. The two agents belong to different factions. \textbf{b) Agent Goal:} their objective is to destroy the opposing faction's base while protecting their own base from being destroyed. \textbf{c) Change:} The two agents have a competitive relationship and need to defeat their opponents.

\subsection{Level 3: Multi-Agent Dynamics} 
Compared to Level 2 (double-agent level), we further increase the number of agents and consider more complex cooperative and competition scenarios in Figure~\ref{fig:combined2}, such as managing dynamic cooperative relationships and facing multiple competition factions.

\vpara{Stage 5: Static Cooperation Competition.} \textbf{a) Agent Setting:} There are four agents and two bases. Four agents are divided into two teams: the Red Team and the Blue Team, with two agents on each team.
\textbf{b) Agent Goal:} The goal is to destroy the opposing faction's base while protecting their own base from being destroyed. \textbf{c) Change:} In this stage, the agents have both cooperative and competitive relationships.

We restrict cooperation to only teammates within the same team, meaning that the agents of the Red Team can cooperate with each other, and the agents of the Blue Team can cooperate with each other. Since the cooperation partners for each agent are fixed and unchanging, we refer to this as static cooperative competition.

\vpara{Stage 6: Dynamic Cooperation Competition.} \textbf{a) Agent Setting:} There are four agents and four bases. Four agents belong to different teams: the Red Team, Blue Team, Green Team, and Yellow Team. \textbf{b) Agent Goal:} Each team can only win by defeating the other three teams. \textbf{c) Change:} This stage introduces a dynamic cooperative relationship.

During the confrontation, in pursuit of ultimate victory, each agent can freely engage in temporary cooperation with one of the other three teams. For example, two teams may choose to collaborate to eliminate a common enemy before turning against each other. Since the cooperation relationships between agents are dynamically changing, we refer to this as dynamic cooperative competition.

\vpara{Stage 7: Hybrid Cooperation Competition.} This stage includes dynamic and static. \textbf{a) Agent Setting:} There are eight agents and four bases. We divide eight agents into four teams, the Red Team, Blue Team, Green Team, and Yellow Team, each team consisting of two agents. \textbf{b) Agent Goal:} Each team can only win by defeating the other three teams. \textbf{c) Change:} This level has both static cooperative relationships and dynamic cooperative relationships.

During the confrontation, the cooperation among teammates remains constant, and the ultimate goal is also aligned: to eliminate the other teams. At the same time, agents can temporarily collaborate with an agent from another team. Because the ultimate goals of the other teams differ, this cooperation relationship is temporary. Therefore, we refer to this scenario as mixed cooperative competition.

\section{Evaluation of \model}

\subsection{Evaluation Setup}

\vpara{Metrics.} In Stage 1 and Stage 2, we primarily use the \textbf{Forward Distance} (F Dis) as the evaluation metric, since the goal of these two stages is to reach the target base. The distance the agent moves toward the base is a suitable metric. Specifically, given the tank's initial position $p_s$, the base's position $p_{target}$, and the tank's position at the end of the game $p_e$, we can calculate the tank's forward distance as follows.
\begin{equation}
    \text{F Dis} = \text{L1}(p_s-p_{target})-\text{L1}(p_e-p_{target}),
\end{equation}
where L1 denotes L1 distance. We also introduced two auxiliary metrics: \textbf{Format Accuracy} (F Acc) and \textbf{Move Accuracy} (M Acc). We predefined the output format for instructions and used regular expressions to match LLM's output. The format accuracy measures the LLM's ability to output in the specified format.
\begin{equation}
    \text{F Acc} = N_{format} / N_{total}, 
\end{equation}
where $N_{format}$ represents the number of turns where the model output is correctly formatted, while $N_{total}$ represents the total number of turns.
Since the target position is known, we can determine whether the agent's movement direction is correct at each step, so the movement accuracy is calculated as follows:
\begin{equation}
    \text{M Acc} = N_{correct} / N_{format},
\end{equation}
where $N_{correct}$ denotes the number of turns with correct movement direction.

In the subsequent five stages, we use \textbf{Score} as the evaluation metric. The agents will earn a certain score for successfully attacking bases, other agents, and NPC tanks. The sum of the three types of scores will serve as the final score. Specifically, the agent's health is 5, and the NPC tank's health is 1. Each attack on the agent and tank obtains 1 score, while each attack on the base obtains 5 scores.

It should be noted that these 5 stages have multiple agents. Each agent needs to connect to a language model. To facilitate the evaluation of different language models, we have adopted the concept of primary and secondary agents. The primary agent is the first agent or the first team of agents, while the remaining agents are secondary agents. Each stage can connect up to two types of language models, with the primary agent connecting to the language model being evaluated, and the secondary agents connecting to another reference language model. Here, we have used yi-1.5-9b as the reference language model. 
By keeping the secondary agents connected to the reference language model unchanged, we can fairly connect and evaluate different language models in the primary agent. For example, in Level 2, there is only one team of 2 agents, so both agents are primary agents, connecting to the language model being evaluated, and their scores are summed for the final score. 

\begin{table}[t!]
\resizebox{0.5\textwidth}{!}{%
\begin{tabular}{@{}cccccccc@{}}
\toprule
\multirow{2}{*}{Models} & \multicolumn{3}{c}{Stage 1} & \multicolumn{3}{c}{Stage 2} & \multirow{2}{*}{Avg. Dis} \\ \cmidrule(lr){2-7}
                        & F Dis   & F Acc   & M Acc  & F Dis   & F Acc   & M Acc  &                           \\ \midrule
\multicolumn{8}{c}{{\textit{API-based models}}}    \\
\cmidrule(lr){1-8}
claude3.5-sonnet-0620   & \textbf{13.7}       & \textbf{1.00}     & 0.97  & \textbf{12.0}       & \textbf{1.00}     & \textbf{0.90}  & \textbf{12.8}                      \\
gpt-4o-mini             & 12.3       & \textbf{1.00}     & \textbf{0.98}  & 10.7       & \textbf{1.00}     & 0.86  & 11.5                      \\
gpt-3.5-turbo-0125      & 10.7       & \textbf{1.00}     & 0.83  & 5.7        & \textbf{1.00}     & 0.65  & 8.2                       \\
glm-4-flash             & 4.3        & \textbf{1.00}     & 0.60  & 5.0        & 0.92     & 0.68  & 4.7                       \\ \midrule
\multicolumn{8}{c}{{\textit{open-source models}}}      \\
\cmidrule(lr){1-8}
internlm2.5-7b-chat     & \textbf{6.7}        & 0.99     & \textbf{0.63}  & 4.7        & 0.99     & 0.63  & \textbf{5.7}                       \\
mistral-7b-instruct     & 3.3        & 0.98     & 0.58  & \textbf{5.8}        & 0.98     & \textbf{0.65}  & 4.6                       \\
glm-4-9b-chat           & 3.8        & \textbf{1.00}     & 0.61  & 4.2        & \textbf{1.00}     & 0.63  & 4.0                       \\
qwen2-7b-Instruct       & 3.7        & 0.98     & 0.60  & 2.1        & 0.99     & 0.56  & 2.9                       \\
yi-1.5-9b-chat-16k      & 1.9        & \textbf{1.00}     & 0.54  & 1.9        & \textbf{1.00}     & 0.55  & 1.9                       \\
gemma2-9b-it            & 2.8        & 0.98     & 0.60  & 0.7        & 0.99     & 0.55  & 1.8                       \\
llama3-8b-instruct      & -1.4       & 0.99     & 0.41  & -1.1       & 0.99     & 0.45  & -1.3                      \\
\cmidrule(lr){1-8}
random                  & 1.0        & -         & 0.49  & 1.4        & -         & 0.52  & 1.2   \\
\bottomrule
\end{tabular}
}
\caption{Evaluation results of Level 1. \textbf{Bold} denotes the best result on API-based models and open-source models.}
\vspace{-4mm}
\label{tab:exp_s1s2}
\end{table}
\begin{table*}[t!]
\centering
\resizebox{0.7\textwidth}{!}{%
\begin{tabular}{@{}cccccccc@{}}
\toprule
\multirow{2}{*}{Models} & \multicolumn{3}{c}{Stage 4}  & \multicolumn{3}{c}{Stage 5}  & \multirow{2}{*}{Avg. Score} \\ \cmidrule(lr){2-7}
                        & Score & F Acc & M Acc & Score & F Acc & M Acc &                             \\ \midrule
\multicolumn{8}{c}{\textit{API-based models}} \\
\cmidrule(lr){1-8}
claude3.5-sonnet-0620   & \textbf{5.3}   & \textbf{1.00}        & \textbf{0.96}      & \textbf{4.3}   & \textbf{1.00}        & \textbf{0.90}      & \textbf{4.8}                         \\
gpt-4o-mini             & 3.3   & 0.96        & 0.69      & 1.3   & 0.90        & 0.65      & 2.3                         \\
gpt-3.5-turbo-0125      & 0.3   & 0.97        & 0.63      & 0.3   & 0.88        & 0.63      & 0.3                         \\
glm-4-flash             & 0.3   & 0.92        & 0.63      & 0.7   & 0.98        & 0.55      & 0.5                         \\ \midrule
\multicolumn{8}{c}{\textit{open-source models}}  \\
\cmidrule(lr){1-8}
internlm2.5-7b-chat     & \textbf{1.8}   & 0.73        & 0.54      & 0.2   & 0.84        & 0.57      & \textbf{1.0}                         \\
yi-1.5-9b-chat-16k      & 1.4   & 0.72        & 0.54      & \textbf{0.4}   & \textbf{0.96}        & 0.46      & 0.9                         \\
llama3-8b-instruct      & 0.4   & 0.81        & 0.34      & \textbf{0.4}   & 0.63        & 0.57      & 0.4                         \\
glm-4-9b-chat           & 0.4   & 0.76        & 0.44      & \textbf{0.4}   & 0.80        & \textbf{0.64}      & 0.4                         \\
qwen2-7b-Instruct       & 0.4   & 0.65        & \textbf{0.58}      & 0.2   & 0.88        & 0.56      & 0.3                         \\
gemma2-9b-it            & 0.4   & \textbf{0.82}        & 0.53      & 0.0   & 0.10        & 0.42      & 0.2                         \\
mistral-7b-instruct     & 0.3   & 0.55        & 0.56      & 0.0   & 0.15        & 0.61      & 0.1                         \\
\cmidrule(lr){1-8}
random                  & 0.2   & -             & 0.48      & 0.0   & -            & 0.49      & 0.1 \\
\bottomrule
\end{tabular}
}
\vspace{-0.2cm}
\caption{Evaluation results of Level 2 of the \model. \textbf{Bold} denotes the best result on API-based models and open-source models.}
\label{tab:exp_s3s4}
\end{table*}

\begin{table*}[t!]
\centering
\resizebox{0.9\textwidth}{!}{%
\begin{tabular}{@{}ccccccccccc@{}}
\toprule
\multirow{2}{*}{Models} & \multicolumn{3}{c}{Stage 5}  & \multicolumn{3}{c}{Stage 6}   & \multicolumn{3}{c}{Stage 7} & \multirow{2}{*}{Avg. Score} \\ 
\cmidrule(lr){2-10}
                        & Score & F Acc & M Acc & Score & F Acc & M Acc & Score & F Acc & M Acc &                             \\ \midrule
\multicolumn{11}{c}{\textit{API-based models}} \\
\cmidrule(lr){1-11}             
claude3.5-sonnet-0620   & \textbf{6.3}   & \textbf{1.00}        & \textbf{0.91}      & \textbf{7.0}   & \textbf{1.00}        & \textbf{0.98}      & \textbf{8.3}   & \textbf{1.00}        & \textbf{0.92}      & \textbf{7.2}                         \\
gpt-4o-mini             & 5.7   & 0.95        & 0.71      & 4.7   & 0.96        & 0.77      & 2.0   & 0.89        & 0.64      & 4.1                         \\
gpt-3.5-turbo-0125      & 1.0   & 0.97        & 0.72      & 1.3   & 0.88        & 0.70      & 2.3   & 0.98        & 0.63      & 1.6                         \\
glm-4-flash             & 0.7   & 0.95        & 0.54      & 2.0   & 0.95        & 0.64      & 0.3   & 0.98        & 0.91      & 1.0                         \\ \midrule
\multicolumn{11}{c}{\textit{open-source models}}                                                                                                                     \\
\cmidrule(lr){1-11}
internlm2.5-7b-chat     & 0.6   & 0.76        & 0.48      & \textbf{2.2}   & 0.67        & 0.42      & \textbf{3.2}   & 0.66        & 0.55      & \textbf{2.0}                         \\
yi-1.5-9b-chat-16k      & 1.0   & \textbf{0.89}        & \textbf{0.56}      & 1.8   & 0.80        & 0.49      & 2.2   & 0.77        & 0.52      & 1.7                         \\
gemma2-9b-it            & \textbf{1.8}   & 0.83        & 0.39      & 0.6   & \textbf{0.87}        & 0.47      & 1.8   & \textbf{0.86}        & 0.57      & 1.4                         \\
llama3-8b-instruct      & 1.0   & 0.85        & 0.47      & 0.8   & 0.85        & 0.55      & 1.2   & 0.83        & \textbf{0.73}      & 1.0                         \\
glm-4-9b-chat           & 1.0   & 0.82        & 0.48      & 0.6   & 0.80        & \textbf{0.63}      & 1.0   & 0.83        & 0.72      & 0.9                         \\
qwen2-7b-Instruct       & 0.0   & 0.69        & 0.52      & 1.4   & 0.63        & 0.47      & 1.2   & 0.65        & 0.64      & 0.9                         \\
mistral-7b-instruct     & 0.2   & 0.49        & 0.38      & 0.0   & 0.54        & 0.43      & 0.2   & 0.47        & 0.49      & 0.1                         \\ 
\cmidrule(lr){1-11}
random                  & 0.2   & -            & 0.50      & 0.4   & -            & 0.49      & 0.6   & -            & 0.52      & 0.4        \\
\bottomrule
\end{tabular}
}
\vspace{-0.2cm}
\caption{Evaluation results of Level 3 of the \model. \textbf{Bold} denotes the best result on API-based models and open-source models.}
\vspace{-4mm}
\label{tab:exp_s5s6s7}
\end{table*}

\vpara{Baselines.} We evaluate two types of models: API-based models and open-source models. The API-based models include Claude3.5\footnote{\textcolor{teal}{https://www.anthropic.com/news/claude-3-5-sonnet}}, GPT-4o-mini~\cite{achiam2023gpt}, GPT-3.5~\cite{achiam2023gpt}, and GLM-4~\cite{glm2024chatglm}. 
The open-source models include Llama3-8b, Mistral-7b~\cite{jiang2023mistral}, Gemma2-9b~\cite{team2024gemma}, GLM-4-9b~\cite{glm2024chatglm}, Yi-1.5-9b~\cite{young2024yi}, Internlm2.5-7b~\cite{team2023internlm}, and Qwen2-7b~\cite{yang2024qwen2}.

\begin{figure}[th!]
    \centering
    \includegraphics[width=0.48\textwidth]{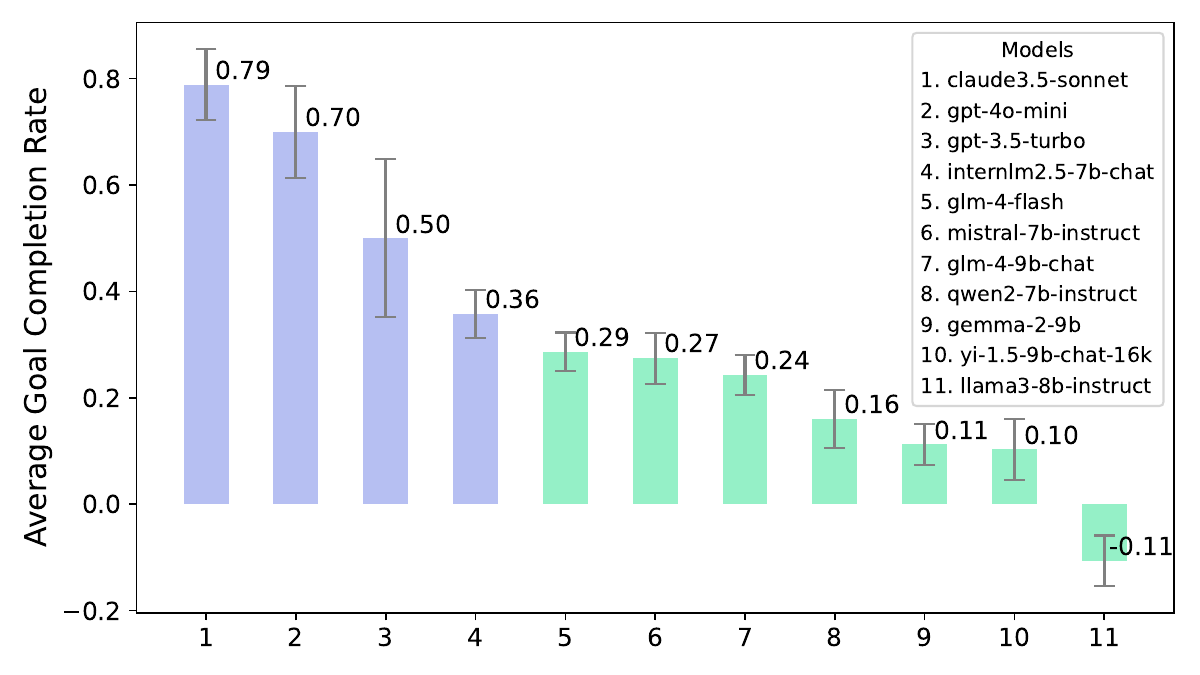}
    \caption{Goal completion rate of different models.} 
    \label{fig:goal_completion}
    \vspace{-4mm}
\end{figure}

\begin{table*}[t!]
\centering
\resizebox{0.9\textwidth}{!}{%
\begin{tabular}{@{}cccccccccccc@{}}
\toprule
\multirow{2}{*}{Models} & \multicolumn{3}{c}{Stage 5}                  & \multicolumn{3}{c}{Stage 6}                  & \multicolumn{3}{c}{Stage 7}                  & \multirow{2}{*}{Avg. Score} & \multirow{2}{*}{$\Delta$ Score} \\ \cmidrule(lr){2-10}
                        & Score        & F Acc         & M Acc         & Score        & F Acc         & M Acc         & Score        & F Acc         & M Acc         &                             &                              \\ \midrule
\multicolumn{12}{c}{\textit{API-based models}}                                                                                                                                                                                             \\ \midrule
claude3.5-sonnet-0620   & \textbf{4.3} & \textbf{1.00} & \textbf{0.87} & \textbf{3.3} & \textbf{1.00} & \textbf{0.87} & \textbf{7.0} & \textbf{1.00} & \textbf{0.89} & \textbf{4.9}                & -2.3                          \\
gpt-4o-mini             & \textbf{4.3} & 0.92          & 0.74          & 3.0          & 0.90          & 0.73          & 4.0          & 0.90          & 0.76          & 3.8                         & -0.3                          \\
gpt-3.5-turbo-0125      & 1.3          & 0.96          & 0.63          & 2.3          & 0.93          & 0.59          & 2.7          & 0.96          & 0.56          & 2.1                         & 0.6                         \\
glm-4-flash             & 1.0          & \textbf{1.00} & 0.48          & 2.0          & \textbf{1.00} & 0.55          & 6.0          & \textbf{1.00} & 0.61          & 3.0                         & 2.0                         \\ \midrule
\multicolumn{12}{c}{\textit{open-source models}}                                                                                                                                                                                           \\ \midrule
yi-1.5-9b-chat-16k      & 1.8          & \textbf{0.94} & 0.56          & 1.6          & \textbf{0.94} & 0.51          & 5.6          & 0.89          & 0.50          & \textbf{3.0}                & 1.3                         \\
gemma2-9b-it            & \textbf{2.0} & \textbf{0.94} & \textbf{0.60} & 2.0          & 0.93          & 0.55          & 4.0          & \textbf{0.92} & \textbf{0.62} & 2.7                         & 1.3                         \\
glm-4-9b-chat           & 0.6          & 0.79          & 0.45          & 1.0          & 0.79          & 0.58          & \textbf{6.0} & 0.82          & 0.57          & 2.5                         & 1.7                         \\
internlm2.5-7b-chat     & 0.6          & 0.78          & 0.45          & \textbf{3.0} & 0.75          & 0.46          & 3.8          & 0.77          & 0.49          & 2.5                         & 0.5                         \\
qwen2-7b-instruct       & 0.0          & 0.81          & 0.50          & 1.0          & 0.85          & 0.57          & 3.6          & 0.80          & 0.55          & 1.5                         & 0.7                         \\
mistral-7b-instruct     & 0.0          & 0.26          & 0.46          & 0.2          & 0.17          & 0.35          & 2.3          & 0.23          & 0.24          & 0.8                         & 0.7                         \\
llama3-8b-instruct      & 0.0          & 0.64          & 0.53          & 1.2          & 0.62          & \textbf{0.69} & 0.8          & 0.55          & 0.55          & 0.7                         & -0.3                          \\ \bottomrule
\end{tabular}
}
\vspace{-0.2cm}
\caption{Ablation Evaluation results of Level 3 of the \model. \textbf{Bold} denotes the best result on API-based models and open-source models. $\Delta$ Score represents Avg. Score in Table 5 subtracts Avg. Score in Table 4.}
\vspace{-4mm}
\label{tab:exp_s5s6s7_ablation}
\end{table*}

\subsection{Overall Experimental Results}

\vpara{Evaluation on Level 1.} As shown in Table \ref{tab:exp_s1s2}, API-based models have a significant advantage over open-source small models in terms of forward distance, especially claude3.5-sonnet and gpt-4o-mini, which can approach the base well. In contrast, open-source models make less progress. Based on the forward distance and total distance to the target, we can calculate the goal completion rate. The results are shown in Figure \ref{fig:goal_completion}. The results show that claude3.5-sonnet and gpt-4o-mini can effectively complete navigation tasks. The goal completion rates of open-source models are all below 50\%.

In addition, open-source small models have lower move accuracy while API-based models have higher move accuracy. This indicates that open-source small models have poor two-dimensional spatial perception, with many movements being incorrect. This explains why these models make less progress. 
Both types of models have high format accuracy, indicating that they can both follow the given action format well.
The evaluation results of these two stages indicate that API-based models can understand the game environment and have a good spatial perception, thus effectively completing objectives. In contrast, open-source models are generally unable to effectively accomplish the goals.

\vpara{Evaluation on Level 2.} From Table \ref{tab:exp_s3s4}, it can be seen that in collaborative and adversarial environments, among all models, only claude3.5-sonnet and gpt-4o-mini achieved relatively high scores, while other models scored comparatively lower. As the difficulty increases, only claude3.5-sonnet maintained a consistent movement accuracy, while the movement accuracy of gpt-4o-mini, gpt-3.5, and glm-4-flash all showed a decline. The movement accuracy of open-source models remains as poor as in Level 1.
Due to the increased complexity of instruction formats, all models experienced some decline in format accuracy. However, API-based models showed a smaller decrease in format accuracy.  These results demonstrate that claude3.5-sonnet has strong environmental understanding, collaboration, and adversarial capabilities, while gpt-4o-mini achieved the second-best performance in these aspects. Other models were unable to cope with complex environments effectively.

\vpara{Evaluation on Level 3.} As shown in Table \ref{tab:exp_s5s6s7}, we can see that in more complex environments, claude3.5-sonnet and gpt-4o-mini still obtain relatively high scores. Other models scored comparatively lower. Notably, only claude3.5-sonnet maintains a high movement accuracy. 
gpt-4o-mini achieved performance second only to claude3.5-sonnet across all metrics. Compared to these two models, other models have lower scores, movement accuracy, and format accuracy. This indicates that these two models have stronger capabilities in dealing with complex environments. The other models perform poorly in lower-level tasks and continue to underperform in higher-level tasks.

\subsection{Ablation Study}

Evaluation results of Level 3 reveal the differences in the comprehensive capabilities of different models in complex environments, including environmental understanding, cooperation, competition abilities, etc.
To assess the effectiveness of models' collaborative abilities,
we conduct further ablation experiments that remove the cooperation interface between agents in the Level 3 evaluation.
This means that while the team relationships between agents remained unchanged, no agent could cooperate with any other. 
The ablation results are shown in Table~\ref{tab:exp_s5s6s7_ablation}.

Comparing Tables \ref{tab:exp_s5s6s7} and Table \ref{tab:exp_s5s6s7_ablation}, we observe that after removing the collaboration interface, only the scores of claude3.5-sonnet and gpt-4o-mini decrease, while the scores of other models increase. This indicates that only claude3.5-sonnet and gpt-4o-mini's cooperation is effective, allowing them to achieve higher scores through cooperation. Conversely, collaboration among other models lowers their scores, indicating that cooperation between these models is ineffective or even harmful. Although claude3.5-sonnet and gpt-4o-mini improve their scores through cooperation, we also notice that the improvement from cooperation is relatively small. Through analysis of the models' battle records, we observe that the models currently exhibit some basic cooperative strategies. We will illustrate this finding through a case study in the following section.

\begin{figure}[t!]
    \centering
    \includegraphics[width=0.45\textwidth]{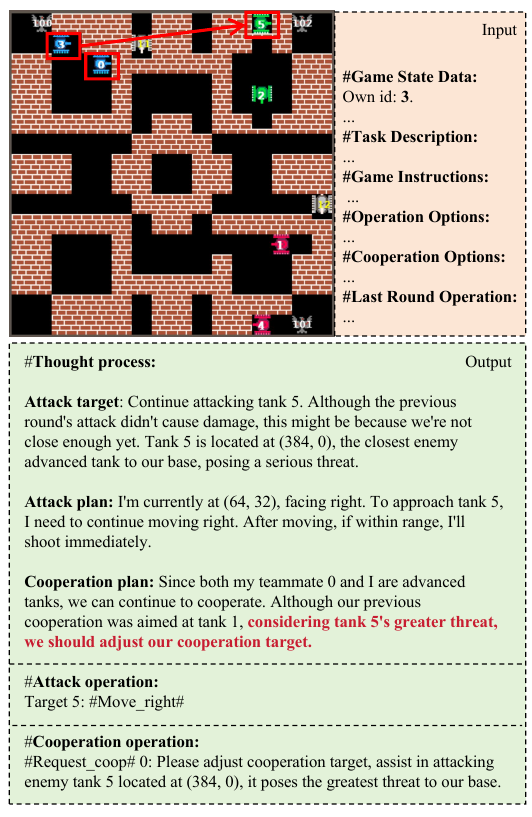}
    \caption{Cooperation example of claude3.5-sonnet to finish decision-making process in Stage 7.} 
    \label{fig:case_study}
    \vspace{-6mm}
\end{figure}

\subsection{Case Study of Level 3 - Stage 7}

In the evaluation of the range from Stage 1 to Stage 7, claude3.5-sonnet shows significant advantages compared to other models. As the game environment becomes increasingly complex, it maintains a stable spatial awareness ability and can move in the correct direction, which is effectively reflected by the M Acc metric. Through analyzing the game confrontation records, we found that claude3.5-sonnet also demonstrated some collaborative abilities. Let's illustrate this with an example in Figure~\ref{fig:case_study}. This example is claude3.5-sonnet's decision-making process in a certain round of Stage 7. In terms of attack operations, claude3.5-sonnet first analyzed the current game situation to identify the biggest threat and moved towards it to prepare for subsequent attacks. In terms of collaborative operations, given the enormous threat posed by this attack target, claude3.5-sonnet timely initiated a collaboration request to teammates to prioritize eliminating this enemy.

\section{Related Works}


\textbf{Multi-Agent Cooperation and Competition.} LLM-based multi-agent focuses on various agent profiles and interactions among agents~\cite{R:1, R:2, R:3}. For instance, CoELA~\cite{R:4} investigated communication and cooperation in LLM-based watch-and-help tasks, while MindAgent~\cite{R:5} assessed the efficiency of collaboration between two or more agents. These efforts primarily focus on multiple agents collaborating effectively to achieve common goals~\cite{R:6, R:35}. In contrast, some efforts focus on the competitive scenario between agents~\cite{R:7}, like AgentPro~\cite{R:8} and ChessGPT~\cite{R:9} to evaluate the strategy reasoning and instruction following the ability of LLM-based agents in competitive settings. Suspicion-Agent~\cite{R:10} utilizes behavioral games theory~\cite{R:30} to evaluate the agent capabilities of LLM in a large set of two-player-two strategies games. However, most of these benchmarks evaluate only a few agents~\cite{R:7}. Instead, our benchmark emphasizes the evaluation of multi-agent coordination and competition.


\noindent \textbf{Benchmarks using Games.} A series of works investigated LLM-based game agents~\cite{R:7, R:11, R:12, R:31}. Several benchmarks center on the game Avalone e.g., AvaloneBench~\cite{R:13}, CodeAct~\cite{R:14}, and Recon~\cite{R:15}. Other benchmarks like CoS~\cite{R:16} and SwarmBrain~\cite{R:17}, which excel in the strategy game StarCraft II~\cite{R:32}. While MindAgent~\cite{R:5}, ProAgent~\cite{R:8}, and S-Agents~\cite{R:18}, among others, focus on Overcooked~\cite{R:19, R:33} and Minecraft~\cite{R:20, R:34} in the context of agent collaboration~\cite{R:21, R:22, R:23}. Moreover, in Minecraft, many efforts ask LLM to generate goals that adapt to the agent’s current state, inventory, acquired skills, and environment~\cite{R:24}. More benchmarks using games include Cradle~\cite{R:25}, Thinker~\cite{R:26}, PokéLLMon~\cite{R:27}, PokerGPT~\cite{R:28}, CreativeAgent~\cite{R:22}, PlayDoom~\cite{R:29}, etc. LLMGA~\cite{R:7} further overviews the benchmarks into six-game genres covering adventure, communication, cooperation, competition, simulation, exploration, and others. However, scarce work investigated multi-agent coordination and competitive scenarios. Even within this narrow field, their focus remains on two-agent interaction with unevenly distributed responsibilities.

\section{Conclusion}

This work introduces the \model, a benchmark for LLM-based multi-agent collaboration and competition evaluation,
which defines seven sub-stages of three varying difficulty levels and conducts a fine-grained evaluation of language models in terms of single-agent scenario navigation capabilities, paired-agent task execution abilities, and multi-agent collaboration and competition capabilities.
We perform a thorough evaluation of 11 LLMs. Although API-based models have demonstrated some collaborative capabilities, there is still enormous room for improvement. 


\bibliography{aaai25}

\UseRawInputEncoding
\onecolumn

\section{Appendix}

\vspace{15pt}

\subsection{Evaluation Setup}

We design 7 stages from easy to difficult for \model. Table \ref{fig:setting} shows the detailed settings for each stage. The setting of the number of turns takes into account the cost factor of calling LLM. Increasing the number of turns may improve the scores of all models. However, the cost of calling the API will also increase accordingly. Therefore, we have set a relatively reasonable turn number for the unified evaluation of all models.

When evaluating each stage, we run multiple times and calculate the average score of the model to assess each model more accurately. For each run, we generate a random map by randomly sampling the agent's position within a small range. Specifically, for API-based models, we run each stage 3 times; for open-source models, we run each stage 5 times.

\begin{table}[h]
\centering
\begin{tabular}{@{}cccccc@{}}
\toprule
Stage & Turns & Agents & Teams & NPC & Goal                            \\ \midrule
1     & 60         & 1            & 1           & 0          & Navigation                      \\
2     & 60         & 1            & 1           & 10         & Navigation                      \\
3     & 80         & 2            & 1           & 10         & Cooperative Task                \\
4     & 80         & 2            & 2           & 10         & Competitive Task                \\
5     & 80         & 4            & 2           & 10         & Static Cooperation Competition  \\
6     & 80         & 4            & 4           & 10         & Dynamic Cooperation Competition \\
7     & 80         & 6            & 3           & 10         & Hybrid Cooperation Competition  \\ \bottomrule
\end{tabular}
\caption{Stage Settings}
\label{fig:setting}
\end{table}

\subsection{Prompt Template}
We use predefined prompt templates to convert game observation data into text prompts, in order to facilitate calling LLM to provide decision results. Figures 1-7 are the prompt templates for each stage. Figure 8-14 shows the Chinese versions of the templates for each stage. Each template consists of 5 parts: game state data, goals, game instructions, operation options, and output format. The main difference is that the goals vary for different stages.

\onecolumn  

\begin{myverbatim}{Prompt Template of Level 1 - Stage 1}
<game>
Current round number: 
%d

Our tank position information (ID, horizontal position, vertical position, direction, health):
%s

Base position information (ID, horizontal position, vertical position):
%s

Map information in front of the tank:
%s

Previous round operation information:
%s
</game>

You are an assistant for a tank battle game, helping users control tanks to achieve victory.
Your ultimate goal is to reach the base on the map in the shortest time possible.

#Game instructions:
- The game map size is 512x512, (0,0) represents the top-left corner, (512,512) represents the bottom-right corner.
- In coordinates (x,y), x represents the horizontal position, y represents the vertical position. Moving left decreases x, moving right increases x, moving up decreases y, moving down increases y.
- The map contains tanks and walls. Tanks are 32x32 in size, walls are 8x8.
- Tanks can move in four directions: up, down, left, and right. Walls and tanks will block tank movement.
- The vertical movement range for tanks is 0-512, the horizontal movement range is 0-512.
- Tanks have 4 directions: up, down, left, and right. Shooting can destroy tanks or walls in front of the current direction.
- When a tank faces the map boundary, it cannot move forward.
- When there's a wall in front of a tank, it needs to shoot to remove the wall before moving forward.

The current game state is given above. Please provide the next action for the tank based on the game state.
You can execute the following defined operations to control the tank.

#Operation options:
- #Move_up#: Move upward
- #Move_down#: Move downward
- #Move_left#: Move left
- #Move_right#: Move right
- #Shoot#: Shoot

#Note
- You can only output one control operation each time.

Your output should follow this format:
#Thought process: 
- Movement plan: {Based on your position, base position, formulate a movement and shooting plan, and decide the next operation}
#Operation: {Specific operation command}
\end{myverbatim}

\newpage

\begin{myverbatim}[break at=23cm]{Prompt Template of Level 1 - Stage 2}
<game>
Current round number: 
%d

Our tank position information (ID, horizontal position, vertical position, direction, health):
%s

Base position information (ID, horizontal position, vertical position):
%s

Enemy tank position information (ID, horizontal position, vertical position, direction, health):
%s

Map information in front of the tank:
%s

Previous round operation information:
%s
</game>

You are an assistant for a tank battle game, helping users control tanks to achieve victory.
Your ultimate goal is to reach the base on the map in the shortest time possible. During movement, you can eliminate enemy tanks that threaten your safety.

#Game instructions:
- The game map size is 512x512, (0,0) represents the top-left corner, (512,512) represents the bottom-right corner.
- In coordinates (x,y), x represents the horizontal position, y represents the vertical position. Moving left decreases x, moving right increases x, moving up decreases y, moving down increases y.
- The map contains tanks and walls. Tanks are 32x32 in size, walls are 8x8.
- Tanks can move in four directions: up, down, left, and right. Walls and tanks will block tank movement.
- The vertical movement range for tanks is 0-512, the horizontal movement range is 0-512.
- Tanks have 4 directions: up, down, left, and right. Shooting can destroy tanks or walls in front of the current direction.
- When a tank faces the map boundary, it cannot move forward.
- When there's a wall in front of a tank, it needs to shoot to remove the wall before moving forward.

The current game state is given above. Please provide the next action for the tank based on the game state.
You can execute the following defined operations to control the tank.

#Operation options:
- #Move_up#: Move upward
- #Move_down#: Move downward
- #Move_left#: Move left
- #Move_right#: Move right
- #Shoot#: Shoot

#Note
- You can only output one control operation each time.

Your output should follow this format:
#Thought process: 
- Movement plan: {Based on your position, base position, and enemy tank positions, formulate a movement and shooting plan, and decide the next operation}
#Operation: {Specific operation command}
\end{myverbatim}

\begin{myverbatim}{Prompt Template of Level 2 - Stage 3}
<game>
Current round number:
%d

Own tank position information (ID, horizontal position, vertical position, direction, health):
%s

Teammate tank position information (ID, horizontal position, vertical position, direction, health):
%s

Our base position information (ID, horizontal position, vertical position):
%s

Enemy base position information (ID, horizontal position, vertical position):
%s

Enemy tank position information (ID, horizontal position, vertical position, direction, health):
%s

Our tanks' attack target information from the last round (our ID, enemy ID):
%s

Historical cooperation attack information:
%s

Map information around the tank:
%s
</game>

You are an assistant for a tank battle game, helping users control tanks to achieve victory.
Your ultimate goal is to destroy the enemy base while protecting our base from being destroyed by the enemy. Destroying enemy tanks also provides rewards. To achieve the ultimate goal, you can cooperate with your teammate.

#Game instructions:
- The game map size is 512x512, (0,0) represents the top-left corner, (512,512) represents the bottom-right corner.
- In coordinates (x,y), x represents the horizontal position, y represents the vertical position. Moving left decreases x, moving right increases x, moving up decreases y, moving down increases y.
- The map contains tanks and walls. Tanks are 32x32 in size, walls are 8x8.
- Tanks can move in four directions: up, down, left, and right. Walls and tanks will block tank movement.
- The vertical movement range for tanks is 0-512, the horizontal movement range is 0-512.
- Tanks have 4 orientations: up, down, left, and right. Shooting can destroy tanks or walls in front of the current direction.
- When a tank faces the map boundary, it cannot move forward.
- When there's a wall in front of a tank, it needs to shoot to remove the wall before moving forward.

The current game state is given above. Please provide the next action for the tank based on the game state.
You can execute the following defined operations to control the tank. You can also choose cooperation options to decide whether to cooperate with teammates.

#Operation options:
- #Move_up#: Move upward
- #Move_down#: Move downward
- #Move_left#: Move left
- #Move_right#: Move right
- #Shoot#: Shoot

#Cooperation options:
- #Request_coop# {Teammate tank ID x}: {Message content}: Send a cooperation message to the tank with ID x
- #Keep_coop#: Maintain cooperation
- #Stop_coop#: Terminate cooperation
- #No_coop#: No cooperation needed

#Note
- When blocked by an enemy tank, shoot immediately to eliminate the enemy.
- For attack effectiveness, don't frequently change attack targets without new emergencies, as this will cause many ineffective movements.
- You can only output one control operation and one cooperation operation each time.

#Last round operation:
- Operation: %s
- Operation feedback: %s

Your output should follow this format:
#Thought process:
- Attack target: {Reason for choosing an attack target, can continue attacking the current target, or choose a new target based on the game state}
- Attack plan: {Based on your position and the attack target's position, make a movement and shooting plan, and decide the next operation}
- Cooperation plan: {Based on your position, teammate's position, and attack target's position, decide on a cooperation plan. You can maintain the previous cooperation plan, initiate a new cooperation request, or terminate the previous cooperation plan}
#Attack operation: Target {Enemy tank ID}: {Specific operation command}
#Cooperation operation: {Specific cooperation command}
\end{myverbatim}

\begin{myverbatim}{Prompt Template of Level 2 - Stage 4}
<game>
Current round number: 
%d

Our tank position information (ID, horizontal position, vertical position, direction, health):
%s

Our base position information (ID, horizontal position, vertical position):
%s

Enemy base position information (ID, horizontal position, vertical position):
%s

Enemy tank position information (ID, horizontal position, vertical position, direction, health):
%s

Map information around the tank:
%s
</game>

You are an assistant for a tank battle game, helping users control tanks to achieve victory.
Your ultimate goal is to destroy the enemy base while protecting our base from being destroyed by the enemy. Destroying enemy tanks also provides rewards.

#Game instructions:
- The game map size is 512x512, (0,0) represents the top-left corner, (512,512) represents the bottom-right corner.
- In coordinates (x,y), x represents the horizontal position, y represents the vertical position. Moving left decreases x, moving right increases x, moving up decreases y, moving down increases y.
- The map contains tanks and walls. Tanks are 32x32 in size, walls are 8x8.
- Tanks can move in four directions: up, down, left, and right. Walls and tanks will block tank movement.
- The vertical movement range for tanks is 0-512, the horizontal movement range is 0-512.
- Tanks have 4 orientations: up, down, left, and right. Shooting can destroy tanks or walls in front of the current direction.
- When a tank faces the map boundary, it cannot move forward.
- When there's a wall in front of a tank, it needs to shoot to remove the wall before moving forward.

The current game state is given above. Please provide the next action for the tank based on the game state.
You can execute the following defined operations to control the tank.

#Operation options:
- #Move_up#: Move upward
- #Move_down#: Move downward
- #Move_left#: Move left
- #Move_right#: Move right
- #Shoot#: Shoot

#Note
- When blocked by an enemy tank, shoot immediately to eliminate the enemy.
- For attack effectiveness, don't frequently change attack targets without new emergencies, as this will cause many ineffective movements.
- You can only output one control operation each time.

#Last round operation:
- Operation: %s
- Operation feedback: %s

Your output should follow this format:
#Thought process: 
- Attack target: {Reason for choosing an attack target, can continue attacking the current target, or choose a new target based on the game state}
- Attack plan: {Based on your position and the attack target's position, make a movement and shooting plan, and decide the next operation}
#Operation: Target {Enemy tank ID}: {Specific operation command}
\end{myverbatim}

\begin{myverbatim}{Prompt Template of Level 3 - Stage 5}
<game>
Current round number:
%d

Own tank position information (ID, horizontal position, vertical position, direction, health):
%s

Teammate tank position information (ID, horizontal position, vertical position, direction, health):
%s

Our base position information (ID, horizontal position, vertical position):
%s

Enemy base position information (ID, horizontal position, vertical position):
%s

Enemy tank position information (ID, horizontal position, vertical position, direction, health):
%s

Our tanks' attack target information from the last round (our ID, enemy ID):
%s

Historical cooperation attack information:
%s

Map information around the tank:
%s
</game>

You are an assistant for a tank battle game, helping users control tanks to achieve victory.
Your ultimate goal is to destroy the enemy base while protecting our base from being destroyed by the enemy. Destroying enemy tanks also provides rewards. To achieve the ultimate goal, you can cooperate with your teammate.

#Game instructions:
- The game map size is 512x512, (0,0) represents the top-left corner, (512,512) represents the bottom-right corner.
- In coordinates (x,y), x represents the horizontal position, y represents the vertical position. Moving left decreases x, moving right increases x, moving up decreases y, moving down increases y.
- The map contains tanks and walls. Tanks are 32x32 in size, walls are 8x8.
- Tanks can move in four directions: up, down, left, and right. Walls and tanks will block tank movement.
- The vertical movement range for tanks is 0-512, the horizontal movement range is 0-512.
- Tanks have 4 orientations: up, down, left, and right. Shooting can destroy tanks or walls in front of the current direction.
- When a tank faces the map boundary, it cannot move forward.
- When there's a wall in front of a tank, it needs to shoot to remove the wall before moving forward.

The current game state is given above. Please provide the next action for the tank based on the game state.
You can execute the following defined operations to control the tank. You can also choose cooperation options to decide whether to cooperate with teammates.

#Operation options:
- #Move_up#: Move upward
- #Move_down#: Move downward
- #Move_left#: Move left
- #Move_right#: Move right
- #Shoot#: Shoot

#Cooperation options:
- #Request_coop# {Teammate tank ID x}: {Message content}: Send a cooperation message to the tank with ID x
- #Keep_coop#: Maintain cooperation
- #Stop_coop#: Terminate cooperation
- #No_coop#: No cooperation needed

#Note
- When blocked by an enemy tank, shoot immediately to eliminate the enemy.
- For attack effectiveness, don't frequently change attack targets without new emergencies, as this will cause many ineffective movements.
- You can only output one control operation and one cooperation operation each time.

#Last round operation:
- Operation: %s
- Operation feedback: %s

Your output should follow this format:
#Thought process:
- Attack target: {Reason for choosing an attack target, can continue attacking the current target, or choose a new target based on the game state}
- Attack plan: {Based on your position and the attack target's position, make a movement and shooting plan, and decide the next operation}
- Cooperation plan: {Based on your position, teammate's position, and attack target's position, decide on a cooperation plan. You can maintain the previous cooperation plan, initiate a new cooperation request, or terminate the previous cooperation plan}
#Attack operation: Target {Enemy tank ID}: {Specific operation command}
#Cooperation operation: {Specific cooperation command}
\end{myverbatim}

\begin{myverbatim}{Prompt Template of Level 3 - Stage 6}
<game>
Current round number:
%d

Our tank position information (ID, horizontal position, vertical position, direction, health, type):
%s

Our base position information (ID, horizontal position, vertical position):
%s

Enemy base position information (ID, horizontal position, vertical position):
%s

Enemy tank position information (ID, horizontal position, vertical position, direction, health, type):
%s

Our tanks' attack target information from the last round (our ID, enemy ID):
%s

Historical cooperation attack information:
%s

Map information around the tank:
%s
</game>

You are an assistant for a tank battle game, helping users control tanks to achieve victory.
Your ultimate goal is to destroy the enemy base while protecting our base from being destroyed by the enemy. Destroying enemy tanks also provides rewards. To achieve the ultimate goal, you can cooperate with an enemy to eliminate other enemies.

#Game instructions:
- The game map size is 512x512, (0,0) represents the top-left corner, (512,512) represents the bottom-right corner.
- In coordinates (x,y), x represents the horizontal position, y represents the vertical position. Moving left decreases x, moving right increases x, moving up decreases y, moving down increases y.
- The map contains tanks and walls. Tanks are 32x32 in size, walls are 8x8.
- Tanks can move in four directions: up, down, left, and right. Walls and tanks will block tank movement.
- The vertical movement range for tanks is 0-512, the horizontal movement range is 0-512.
- Tanks have 4 orientations: up, down, left, and right. Shooting can destroy tanks or walls in front of the current direction.
- When a tank faces the map boundary, it cannot move forward.
- When there's a wall in front of a tank, it needs to shoot to remove the wall before moving forward.
- Tanks have two types: normal and advanced. Only advanced tanks have cooperation capabilities.

The current game state is given above. Please provide the next action for the tank based on the game state.
You can execute the following defined operations to control the tank. You can also choose cooperation options to decide whether to cooperate with teammates.

#Operation options:
- #Move_up#: Move upward
- #Move_down#: Move downward
- #Move_left#: Move left
- #Move_right#: Move right
- #Shoot#: Shoot

#Cooperation options:
- #Request_coop# {Tank ID x}: {Message content}: Send a cooperation message to the tank with ID x
- #Keep_coop#: Maintain cooperation
- #Stop_coop#: Terminate cooperation
- #No_coop#: No cooperation needed

#Note
- When blocked by an enemy tank, shoot immediately to eliminate the enemy.
- For attack effectiveness, don't frequently change attack targets without new emergencies, as this will cause many ineffective movements.
- You can only output one control operation and one cooperation operation each time.

#Last round operation:
- Operation: %s
- Operation feedback: %s

Your output should follow this format:
#Thought process:
- Attack target: {Reason for choosing an attack target, can continue attacking the current target, or choose a new target based on the game state}
- Attack plan: {Based on your position and the attack target's position, make a movement and shooting plan, and decide the next operation}
- Cooperation plan: {Based on your position and the attack target's position, decide on a cooperation plan. You can maintain the previous round's cooperation plan, initiate a new cooperation request, or terminate the previous round's cooperation plan}
#Attack operation: Target {Enemy tank ID}: {Specific operation command}
#Cooperation operation: {Specific cooperation command}
\end{myverbatim}

\begin{myverbatim}{Prompt Template of Level 3 - Stage 7}
<game>
Current round number: 
%d

Own tank position information (ID, horizontal position, vertical position, direction, health, type):
%s

Teammate tank position information (ID, horizontal position, vertical position, direction, health, type):
%s

Our base position information (ID, horizontal position, vertical position):
%s

Enemy base position information (ID, horizontal position, vertical position):
%s

Enemy tank position information (ID, horizontal position, vertical position, direction, health, type):
%s

Our tanks' attack target information from the last round (our ID, enemy ID):
%s

Historical cooperation attack information:
%s

Map information around the tank:
%s
</game>

You are an assistant for a tank battle game, helping users control tanks to achieve victory.
Your ultimate goal is to destroy the enemy base while protecting our base from being destroyed by the enemy. Destroying enemy tanks also provides rewards. To achieve the ultimate goal, you can cooperate with your teammates, or temporarily cooperate with an enemy to eliminate other enemies.

#Game instructions:
- The game map size is 512x512, (0,0) represents the top-left corner, (512,512) represents the bottom-right corner.
- In coordinates (x,y), x represents the horizontal position, y represents the vertical position. Moving left decreases x, moving right increases x, moving up decreases y, moving down increases y.
- The map contains tanks and walls. Tanks are 32x32 in size, walls are 8x8.
- Tanks can move in four directions: up, down, left, and right. Walls and tanks will block tank movement.
- The vertical movement range for tanks is 0-512, the horizontal movement range is 0-512.
- Tanks have 4 orientations: up, down, left, and right. Shooting can destroy tanks or walls in front of the current direction.
- When a tank faces the map boundary, it cannot move forward.
- When there's a wall in front of a tank, it needs to shoot to remove the wall before moving forward.
- Tanks have two types: normal and advanced. Only advanced tanks have cooperation capabilities.

The current game state is given above. Please provide the next action for the tank based on the game state.
You can execute the following defined operations to control the tank. You can also choose cooperation options to decide whether to cooperate with teammates.

#Operation options:
- #Move_up#: Move upward
- #Move_down#: Move downward
- #Move_left#: Move left
- #Move_right#: Move right
- #Shoot#: Shoot

#Cooperation options:
- #Request_coop# {Tank ID x}: {Message content}: Send a cooperation message to the tank with ID x
- #Keep_coop#: Maintain cooperation
- #Stop_coop#: Terminate cooperation
- #No_coop#: No cooperation needed

#Note
- When blocked by an enemy tank, shoot immediately to eliminate the enemy.
- For attack effectiveness, don't frequently change attack targets without new emergencies, as this will cause many ineffective movements.
- You can only output one control operation and one cooperation operation each time.

#Last round operation:
- Operation: %s
- Operation feedback: %s

Your output should follow this format:
#Thought process:
- Attack target: {Reason for choosing an attack target, can continue attacking the current target, or choose a new target based on the game state}
- Attack plan: {Based on your position and the attack target's position, make a movement and shooting plan, and decide the next operation}
- Cooperation plan: {Based on your position and the attack target's position, decide on a cooperation plan. You can maintain the previous round's cooperation plan, initiate a new cooperation request, or terminate the previous round's cooperation plan}
#Attack operation: Target {Enemy tank ID}: {Specific operation command}
#Cooperation operation: {Specific cooperation command}
\end{myverbatim}














\begin{CJK*}{UTF8}{gbsn}
\begin{myverbatim3}{Prompt Template of Level 1 - Stage 1 in Chinese}
|\begin{CJK*}{UTF8}{gbsn}
\begin{Verbatim}[numbers=left, numbersep=10pt,baselinestretch=1.3,breaklines=true, formatcom=\renewcommand{\theFancyVerbLine}{\fontsize{9}{40}\fontfamily{ptm}\selectfont\arabic{FancyVerbLine}}]
<game>
当前会合数: 
%d

我方坦克位置信息（编号，水平位置，垂直位置，朝向，血量）:
%s

基地位置信息（编号，水平位置，垂直位置）:
%s

坦克前方地图信息:
%s

上一会合操作信息:

</game>

你是一个坦克对战游戏的助手，可以帮助用户在游戏中控制坦克取得胜利。
你的最终目标是以最短时间到达地图中的基地。

#游戏说明：
- 游戏地图大小为512x512，(0，0)表示左上角，(512，512)表示右下角。
- 坐标(x，y)中x表示水平位置，y表示垂直位置，向左移动x减小，向右移动x增大，向上移动y减小，向下移动y增大。
- 地图中有坦克及wall，坦克大小为32x32，wall大小为8x8。
- 坦克可以向上、下、左、右四个方向移动，wall和坦克会阻挡坦克的移动。
- 坦克上下的移动范围为0-512，左右的移动范围0-512。
- 坦克有4个朝向，上、下、左、右，射击可以销毁当前方向前方的坦克或wall。
- 坦克前方为地图边界时，无法继续向前移动。
- 坦克前方有wall时，需要射击消除wall之后才能继续向前移动。

上面给出了当前游戏的状态，请根据游戏状态，给出坦克的下一步操作。
你可以执行以下定义的操作来控制坦克。

#操作选项:
- #Move_up#: 向上移动
- #Move_down#: 向下移动
- #Move_left#: 向左移动
- #Move_right#: 向右移动
- #Shoot#: 射击

#注意
- 你每次只能输出一个操作。

你的输出应该符合以下格式:
#思考过程: 
- 移动计划: {根据自己的位置和基地位置，制订移动和射击计划，并决定下一步的操作}
#操作: {具体操作指令}
\end{Verbatim}
\end{CJK*}|
\end{myverbatim3}
\end{CJK*}

\begin{CJK*}{UTF8}{gbsn}
\begin{myverbatim3}{Prompt Template of Level 1 - Stage 2 in Chinese}
|\begin{CJK*}{UTF8}{gbsn}
\begin{Verbatim}[numbers=left, numbersep=10pt,baselinestretch=1.3,breaklines=true, formatcom=\renewcommand{\theFancyVerbLine}{\fontsize{9}{40}\fontfamily{ptm}\selectfont\arabic{FancyVerbLine}}]
<game>
当前会合数: 
%d

我方坦克位置信息（编号，水平位置，垂直位置，朝向，血量）:
%s

基地位置信息（编号，水平位置，垂直位置）:
%s

敌方坦克位置信息（编号，水平位置，垂直位置，朝向，血量）:
%s

坦克前方地图信息:
%s

上一会合操作信息:
%s
</game>

你是一个坦克对战游戏的助手，可以帮助用户在游戏中控制坦克取得胜利。
你的最终目标是以最短时间到达地图中的基地。移动过程中你可以消灭影响你安全的敌方坦克。

#游戏说明：
- 游戏地图大小为512x512，(0，0)表示左上角，（512，512）表示右下角。
- 坐标（x，y）中x表示水平位置，y表示垂直位置，向左移动x减小，向右移动x增大，向上移动y减小，向下移动y增大。
- 地图中有坦克及wall，坦克大小为32x32，wall大小为8x8。
- 坦克可以向上、下、左、右四个方向移动，wall和坦克会阻挡坦克的移动。
- 坦克上下的移动范围为0-512，左右的移动范围0-512。
- 坦克有4个朝向，上、下、左、右，射击可以销毁当前方向前方的坦克或wall。
- 坦克前方为地图边界时，无法继续向前移动。
- 坦克前方有wall时，需要射击消除wall之后才能继续向前移动。

上面给出了当前游戏的状态，请根据游戏状态，给出坦克的下一步操作。
你可以执行以下定义的操作来控制坦克。

#操作选项:
- #Move_up#: 向上移动
- #Move_down#: 向下移动
- #Move_left#: 向左移动
- #Move_right#: 向右移动
- #Shoot#: 射击

#注意
- 你每次只能输出一个操作。

你的输出应该符合以下格式:
#思考过程: 
- 移动计划: {根据自己的位置和基地位置，以及敌方坦克的位置，制订移动和射击计划，并决定下一步的操作}
#操作: {具体操作指令}
\end{Verbatim}
\end{CJK*}|
\end{myverbatim3}
\end{CJK*}

\begin{CJK*}{UTF8}{gbsn}
\begin{myverbatim3}{Prompt Template of Level 2 - Stage 3 in Chinese}
|\begin{CJK*}{UTF8}{gbsn}
\begin{Verbatim}[numbers=left, numbersep=10pt,baselinestretch=1.3,breaklines=true,formatcom=\renewcommand{\theFancyVerbLine}{\fontsize{9}{40}\fontfamily{ptm}\selectfont\arabic{FancyVerbLine}}]
<game>
当前回合数: 
%d

自己坦克位置信息（编号，水平位置，垂直位置，朝向，血量）:
%s

队友坦克位置信息（编号，水平位置，垂直位置，朝向，血量）:
%s

我方基地位置信息（编号，水平位置，垂直位置）:
%s

敌方基地位置信息（编号，水平位置，垂直位置）:
%s

敌方坦克位置信息（编号，水平位置，垂直位置，朝向，血量）:
%s

我方坦克上一回合攻击目标信息（我方编号，敌方编号）:
%s

历史协作攻击信息:
%s

坦克周围地图信息:
%s
</game>

你是一个坦克对战游戏的助手，可以帮助用户在游戏中控制坦克取得胜利。
你的最终目标是消灭敌方基地，并保护我方基地不被敌人消灭。消灭敌方坦克也有奖励。你可以与你的队友协作完成目标。

#游戏说明：
- 游戏地图大小为512x512，(0，0)表示左上角，（512，512）表示右下角。
- 坐标（x，y）中x表示水平位置，y表示垂直位置，向左移动x减小，向右移动x增大，向上移动y减小，向下移动y增大。
- 地图中有坦克及wall，坦克大小为32x32，wall大小为8x8。
- 坦克可以向上、下、左、右四个方向移动，wall和坦克会阻挡坦克的移动。
- 坦克上下的移动范围为0-512，左右的移动范围0-512。
- 坦克有4个朝向，上、下、左、右，射击可以销毁当前方向前方的坦克或wall。
- 坦克前方为地图边界时，无法继续向前移动。
- 坦克前方有wall时，需要射击消除wall之后才能继续向前移动。

上面给出了当前游戏的状态，请根据游戏状态，给出坦克的下一步操作。
你可以执行以下定义的操作来控制坦克。并可以选择协作选项决定是否与队友协作攻击。

#操作选项:
- #Move_up#: 向上移动
- #Move_down#: 向下移动
- #Move_left#: 向左移动
- #Move_right#: 向右移动
- #Shoot#: 射击

#协作选项:
- #Request_coop# {队友坦克编号x}: {消息内容}: 向编号为x的坦克发送协作消息
- #Keep_coop#: 保持协作
- #Stop_coop#: 终止协作
- #No_coop#: 无需协作

#注意
- 当阻挡自己的为敌方坦克时，立即射击消灭敌方。
- 为了攻击的有效性，没有新的突发情况，不要频繁地更换攻击目标，这样会造成很多无效移动。
- 你每次只能输出一个控制操作和一个协作操作。

#上一回合操作: 
- 操作: %s
- 操作反馈: %s

你的输出应该符合以下格式:
#思考过程: 
- 攻击目标: {选择某个攻击目标的原因，可以继续攻击当前目标，或者根据游戏状态选择新的攻击目标}
- 攻击计划: {根据自己的位置和攻击目标的位置，制订移动和射击计划，并决定下一步的操作}
- 协作计划: {根据自己的位置、队友的位置和攻击目标的位置，决定协作计划，可以保持上一会合的协作计划，也可发起新的协作请求，也可终止上一会合协作计划}
#攻击操作: Target {敌方坦克编号}: {具体操作指令}
#协作操作: {具体协作指令}
\end{Verbatim}
\end{CJK*}|
\end{myverbatim3}
\end{CJK*}

\begin{CJK*}{UTF8}{gbsn}
\begin{myverbatim3}{Prompt Template of Level 2 - Stage 4 in Chinese}
|\begin{CJK*}{UTF8}{gbsn}
\begin{Verbatim}[numbers=left, numbersep=10pt,baselinestretch=1.3,breaklines=true, formatcom=\renewcommand{\theFancyVerbLine}{\fontsize{9}{40}\fontfamily{ptm}\selectfont\arabic{FancyVerbLine}}]
<game>
当前回合数: 
%d

我方坦克位置信息（编号，水平位置，垂直位置，朝向，血量）:
%s

我方基地位置信息（编号，水平位置，垂直位置）:
%s

敌方基地位置信息（编号，水平位置，垂直位置）:
%s

敌方坦克位置信息（编号，水平位置，垂直位置，朝向，血量）:
%s

坦克周围地图信息:
%s
</game>

你是一个坦克对战游戏的助手，可以帮助用户在游戏中控制坦克取得胜利。
你的最终目标是消灭敌方基地，并保护我方基地不被敌人消灭。消灭敌方坦克也有奖励。

#游戏说明：
- 游戏地图大小为512x512，(0，0)表示左上角，（512，512）表示右下角。
- 坐标（x，y）中x表示水平位置，y表示垂直位置，向左移动x减小，向右移动x增大，向上移动y减小，向下移动y增大。
- 地图中有坦克及wall，坦克大小为32x32，wall大小为8x8。
- 坦克可以向上、下、左、右四个方向移动，wall和坦克会阻挡坦克的移动。
- 坦克上下的移动范围为0-512，左右的移动范围0-512。
- 坦克有4个朝向，上、下、左、右，射击可以销毁当前方向前方的坦克或wall。
- 坦克前方为地图边界时，无法继续向前移动。
- 坦克前方有wall时，需要射击消除wall之后才能继续向前移动。

上面给出了当前游戏的状态，请根据游戏状态，给出坦克的下一步操作。
你可以执行以下定义的操作来控制坦克。

#操作选项:
- #Move_up#: 向上移动
- #Move_down#: 向下移动
- #Move_left#: 向左移动
- #Move_right#: 向右移动
- #Shoot#: 射击

#注意
- 当阻挡自己的为敌方坦克时，立即射击消灭敌方。
- 为了攻击的有效性，没有新的突发情况，不要频繁地更换攻击目标，这样会造成很多无效移动。
- 你每次只能输出一个操作。

#上一回合操作: 
- 操作: %s
- 操作反馈: %s

你的输出应该符合以下格式:
#思考过程: 
- 攻击目标: {选择某个攻击目标的原因，可以继续攻击当前目标，或者根据游戏状态选择新的攻击目标}
- 攻击计划: {根据自己的位置和攻击目标的位置，制订移动和射击计划，并决定下一步的操作}
#操作: Target {敌方坦克编号}: {具体操作指令}
\end{Verbatim}
\end{CJK*}|
\end{myverbatim3}
\end{CJK*}

\begin{CJK*}{UTF8}{gbsn}
\begin{myverbatim3}{Prompt Template of Level 3 - Stage 5 in Chinese}
|\begin{CJK*}{UTF8}{gbsn}
\begin{Verbatim}[numbers=left, numbersep=10pt,baselinestretch=1.3,breaklines=true, formatcom=\renewcommand{\theFancyVerbLine}{\fontsize{9}{40}\fontfamily{ptm}\selectfont\arabic{FancyVerbLine}}]
<game>
当前回合数: 
%d

自己坦克位置信息（编号，水平位置，垂直位置，朝向，血量）:
%s

队友坦克位置信息（编号，水平位置，垂直位置，朝向，血量）:
%s

我方基地位置信息（编号，水平位置，垂直位置）:
%s

敌方基地位置信息（编号，水平位置，垂直位置）:
%s

敌方坦克位置信息（编号，水平位置，垂直位置，朝向，血量）:
%s

我方坦克上一回合攻击目标信息（我方编号，敌方编号）:
%s

历史协作攻击信息:
%s

坦克周围地图信息:
%s
</game>

你是一个坦克对战游戏的助手，可以帮助用户在游戏中控制坦克取得胜利。
你的最终目标是消灭敌方基地，并保护我方基地不被敌人消灭。消灭敌方坦克也有奖励。你可以与你的队友协作完成目标。

#游戏说明：
- 游戏地图大小为512x512，(0，0)表示左上角，（512，512）表示右下角。
- 坐标（x，y）中x表示水平位置，y表示垂直位置，向左移动x减小，向右移动x增大，向上移动y减小，向下移动y增大。
- 地图中有坦克及wall，坦克大小为32x32，wall大小为8x8。
- 坦克可以向上、下、左、右四个方向移动，wall和坦克会阻挡坦克的移动。
- 坦克上下的移动范围为0-512，左右的移动范围0-512。
- 坦克有4个朝向，上、下、左、右，射击可以销毁当前方向前方的坦克或wall。
- 坦克前方为地图边界时，无法继续向前移动。
- 坦克前方有wall时，需要射击消除wall之后才能继续向前移动。

上面给出了当前游戏的状态，请根据游戏状态，给出坦克的下一步操作。
你可以执行以下定义的操作来控制坦克。并可以选择协作选项决定是否与队友协作攻击。

#操作选项:
- #Move_up#: 向上移动
- #Move_down#: 向下移动
- #Move_left#: 向左移动
- #Move_right#: 向右移动
- #Shoot#: 射击

#协作选项:
- #Request_coop# {队友坦克编号x}: {消息内容}: 向编号为x的坦克发送协作消息
- #Keep_coop#: 保持协作
- #Stop_coop#: 终止协作
- #No_coop#: 无需协作

#注意
- 当阻挡自己的为敌方坦克时，立即射击消灭敌方。
- 为了攻击的有效性，没有新的突发情况，不要频繁地更换攻击目标，这样会造成很多无效移动。
- 你每次只能输出一个控制操作和一个协作操作。

#上一回合操作: 
- 操作: %s
- 操作反馈: %s

你的输出应该符合以下格式:
#思考过程: 
- 攻击目标: {选择某个攻击目标的原因，可以继续攻击当前目标，或者根据游戏状态选择新的攻击目标}
- 攻击计划: {根据自己的位置和攻击目标的位置，制订移动和射击计划，并决定下一步的操作}
- 协作计划: {根据自己的位置、队友的位置和攻击目标的位置，决定协作计划，可以保持上一会合的协作计划，也可发起新的协作请求，也可终止上一会合协作计划}
#攻击操作: Target {敌方坦克编号}: {具体操作指令}
#协作操作: {具体协作指令}
\end{Verbatim}
\end{CJK*}|
\end{myverbatim3}
\end{CJK*}

\begin{CJK*}{UTF8}{gbsn}
\begin{myverbatim3}{Prompt Template of Level 3 - Stage 6 in Chinese}
|\begin{CJK*}{UTF8}{gbsn}
\begin{Verbatim}[numbers=left, numbersep=10pt,baselinestretch=1.3,breaklines=true, formatcom=\renewcommand{\theFancyVerbLine}{\fontsize{9}{40}\fontfamily{ptm}\selectfont\arabic{FancyVerbLine}}]
<game>
当前回合数: 
%d

我方坦克位置信息（编号，水平位置，垂直位置，朝向，血量，类型）:
%s

我方基地位置信息（编号，水平位置，垂直位置）:
%s

敌方基地位置信息（编号，水平位置，垂直位置）:
%s

敌方坦克位置信息（编号，水平位置，垂直位置，朝向，血量，类型）:
%s

我方坦克上一回合攻击目标信息（我方编号，敌方编号）:
%s

历史协作攻击信息:
%s

坦克周围地图信息:
%s
</game>

你是一个坦克对战游戏的助手，可以帮助用户在游戏中控制坦克取得胜利。
你的最终目标是消灭敌方基地，并保护我方基地不被敌人消灭。消灭敌方坦克也有奖励。为了完成最终目标，你可以与某个敌人协作消灭其它敌人。

#游戏说明：
- 游戏地图大小为512x512，(0，0)表示左上角，（512，512）表示右下角。
- 坐标（x，y）中x表示水平位置，y表示垂直位置，向左移动x减小，向右移动x增大，向上移动y减小，向下移动y增大。
- 地图中有坦克及wall，坦克大小为32x32，wall大小为8x8。
- 坦克可以向上、下、左、右四个方向移动，wall和坦克会阻挡坦克的移动。
- 坦克上下的移动范围为0-512，左右的移动范围0-512。
- 坦克有4个朝向，上、下、左、右，射击可以销毁当前方向前方的坦克或wall。
- 坦克前方为地图边界时，无法继续向前移动。
- 坦克前方有wall时，需要射击消除wall之后才能继续向前移动。
- 坦克有普通和高级两种类型，只有高级坦克具有协作能力。

上面给出了当前游戏的状态，请根据游戏状态，给出坦克的下一步操作。
你可以执行以下定义的操作来控制坦克。并可以选择协作选项决定是否与队友协作攻击。

#操作选项:
- #Move_up#: 向上移动
- #Move_down#: 向下移动
- #Move_left#: 向左移动
- #Move_right#: 向右移动
- #Shoot#: 射击

#协作选项:
- #Request_coop# {坦克编号x}: {消息内容}: 向编号为x的坦克发送协作消息
- #Keep_coop#: 保持协作
- #Stop_coop#: 终止协作
- #No_coop#: 无需协作

#注意
- 当阻挡自己的为敌方坦克时，立即射击消灭敌方。
- 为了攻击的有效性，没有新的突发情况，不要频繁地更换攻击目标，这样会造成很多无效移动。
- 你每次只能输出一个控制操作和一个协作操作。

#上一回合操作: 
- 操作: %s
- 操作反馈: %s

你的输出应该符合以下格式:
#思考过程: 
- 攻击目标: {选择某个攻击目标的原因，可以继续攻击当前目标，或者根据游戏状态选择新的攻击目标}
- 攻击计划: {根据自己的位置和攻击目标的位置，制订移动和射击计划，并决定下一步的操作}
- 协作计划: {根据自己的位置、攻击目标的位置，决定协作计划，可以保持上一会合的协作计划，也可发起新的协作请求，也可终止上一会合协作计划}
#攻击操作: Target {敌方坦克编号}: {具体操作指令}
#协作操作: {具体协作指令}
\end{Verbatim}
\end{CJK*}|
\end{myverbatim3}
\end{CJK*}

\begin{CJK*}{UTF8}{gbsn}
\begin{myverbatim3}{Prompt Template of Level 3 - Stage 7 in Chinese}
|\begin{CJK*}{UTF8}{gbsn}
\begin{Verbatim}[numbers=left, numbersep=10pt,baselinestretch=1.3,breaklines=true, formatcom=\renewcommand{\theFancyVerbLine}{\fontsize{9}{40}\fontfamily{ptm}\selectfont\arabic{FancyVerbLine}}]
<game>
当前回合数: 
%d

自己坦克位置信息（编号，水平位置，垂直位置，朝向，血量，类型）:
%s

队友坦克位置信息（编号，水平位置，垂直位置，朝向，血量，类型）:
%s

我方基地位置信息（编号，水平位置，垂直位置）:
%s

敌方基地位置信息（编号，水平位置，垂直位置）:
%s

敌方坦克位置信息（编号，水平位置，垂直位置，朝向，血量，类型）:
%s

我方坦克上一回合攻击目标信息（我方编号，敌方编号）:
%s

历史协作攻击信息:
%s

坦克周围地图信息:
%s
</game>

你是一个坦克对战游戏的助手，可以帮助用户在游戏中控制坦克取得胜利。
你的最终目标是消灭敌方基地，并保护我方基地不被敌人消灭。消灭敌方坦克也有奖励。为了完成最终目标，你可以与你的队友协作，也可以暂时与某个敌人协作消灭其它敌人。

#游戏说明：
- 游戏地图大小为512x512，(0，0)表示左上角，（512，512）表示右下角。
- 坐标（x，y）中x表示水平位置，y表示垂直位置，向左移动x减小，向右移动x增大，向上移动y减小，向下移动y增大。
- 地图中有坦克及wall，坦克大小为32x32，wall大小为8x8。
- 坦克可以向上、下、左、右四个方向移动，wall和坦克会阻挡坦克的移动。
- 坦克上下的移动范围为0-512，左右的移动范围0-512。
- 坦克有4个朝向，上、下、左、右，射击可以销毁当前方向前方的坦克或wall。
- 坦克前方为地图边界时，无法继续向前移动。
- 坦克前方有wall时，需要射击消除wall之后才能继续向前移动。
- 坦克有普通和高级两种类型，只有高级坦克具有协作能力。

上面给出了当前游戏的状态，请根据游戏状态，给出坦克的下一步操作。
你可以执行以下定义的操作来控制坦克。并可以选择协作选项决定是否与队友协作攻击。

#操作选项:
- #Move_up#: 向上移动
- #Move_down#: 向下移动
- #Move_left#: 向左移动
- #Move_right#: 向右移动
- #Shoot#: 射击

#协作选项:
- #Request_coop# {坦克编号x}: {消息内容}: 向编号为x的坦克发送协作消息
- #Keep_coop#: 保持协作
- #Stop_coop#: 终止协作
- #No_coop#: 无需协作

#注意
- 当阻挡自己的为敌方坦克时，立即射击消灭敌方。
- 为了攻击的有效性，没有新的突发情况，不要频繁地更换攻击目标，这样会造成很多无效移动。
- 你每次只能输出一个控制操作和一个协作操作。

#上一回合操作: 
- 操作: %s
- 操作反馈: %s

你的输出应该符合以下格式:
#思考过程: 
- 攻击目标: {选择某个攻击目标的原因，可以继续攻击当前目标，或者根据游戏状态选择新的攻击目标}
- 攻击计划: {根据自己的位置和攻击目标的位置，制订移动和射击计划，并决定下一步的操作}
- 协作计划: {根据自己的位置、攻击目标的位置，决定协作计划，可以保持上一会合的协作计划，也可发起新的协作请求，也可终止上一会合协作计划}
#攻击操作: Target {敌方坦克编号}: {具体操作指令}
#协作操作: {具体协作指令}
\end{Verbatim}
\end{CJK*}|
\end{myverbatim3}
\end{CJK*}


\end{document}